\documentclass[10pt,twocolumn,letterpaper]{article}
\usepackage{xcolor}         
\usepackage{subfigure}
\usepackage{wrapfig}
\usepackage{iccv}
\usepackage{times}
\usepackage{epsfig}
\usepackage{graphicx}
\usepackage{amsmath}
\usepackage{amssymb}

\usepackage{ctable}
\usepackage{color}

\usepackage{caption}
\usepackage{multirow}
\usepackage{float}
\usepackage{stfloats}

\usepackage{stmaryrd}
\usepackage{mmstyle}

\usepackage[pagebackref=true,breaklinks=true,letterpaper=true,colorlinks,bookmarks=false]{hyperref}
\usepackage{color,colortbl}

\definecolor{Gray}{gray}{0.9}
\definecolor{White}{gray}{1}
\iccvfinalcopy 

\title{Learning Skeletal Graph Neural Networks for Hard 3D Pose Estimation}

\newcommand*{\affaddr}[1]{#1} 
\newcommand*{\affmark}[1][*]{\textsuperscript{#1}}

\makeatletter
\newcommand{\printfnsymbol}[1]{%
  \textsuperscript{\@fnsymbol{#1}}%
}
\makeatother
\author{%
Ailing Zeng\affmark[1], Xiao Sun\affmark[2], Lei Yang\affmark[3], Nanxuan Zhao\affmark[1], Minhao Liu\affmark[1], Qiang Xu\affmark[1]\\
\affaddr{\affmark[1]The Chinese University of Hong Kong}\\
\affaddr{\affmark[2]Microsoft Research Asia},
\affaddr{\affmark[3]Sensetime Group Ltd.}\\
}

\begin{document}
\maketitle

\begin{abstract}

Various deep learning techniques have been proposed to solve the single-view 2D-to-3D pose estimation problem. While the average prediction accuracy has been improved significantly over the years, the performance on hard poses with depth ambiguity, self-occlusion, and complex or rare poses is still far from satisfactory. In this work, we target these hard poses and present a novel skeletal GNN learning solution. To be specific, we propose a hop-aware hierarchical channel-squeezing fusion layer to effectively extract relevant information from neighboring nodes while suppressing undesired noises in GNN learning. In addition, we propose a temporal-aware dynamic graph construction procedure that is robust and effective for 3D pose estimation. Experimental results on the Human3.6M dataset show that our solution achieves 10.3\% average prediction accuracy improvement and greatly improves on hard poses over state-of-the-art techniques. We further apply the proposed technique on the skeleton-based action recognition task and also achieve state-of-the-art performance. Our code is available at \url{https://github.com/ailingzengzzz/Skeletal-GNN}.
\end{abstract}
\vspace{-10pt}

\section{Introduction}
\label{sec:intro}

\begin{figure*}
\begin{center}
\vspace{-0.5cm}

\includegraphics[width=0.96\textwidth]{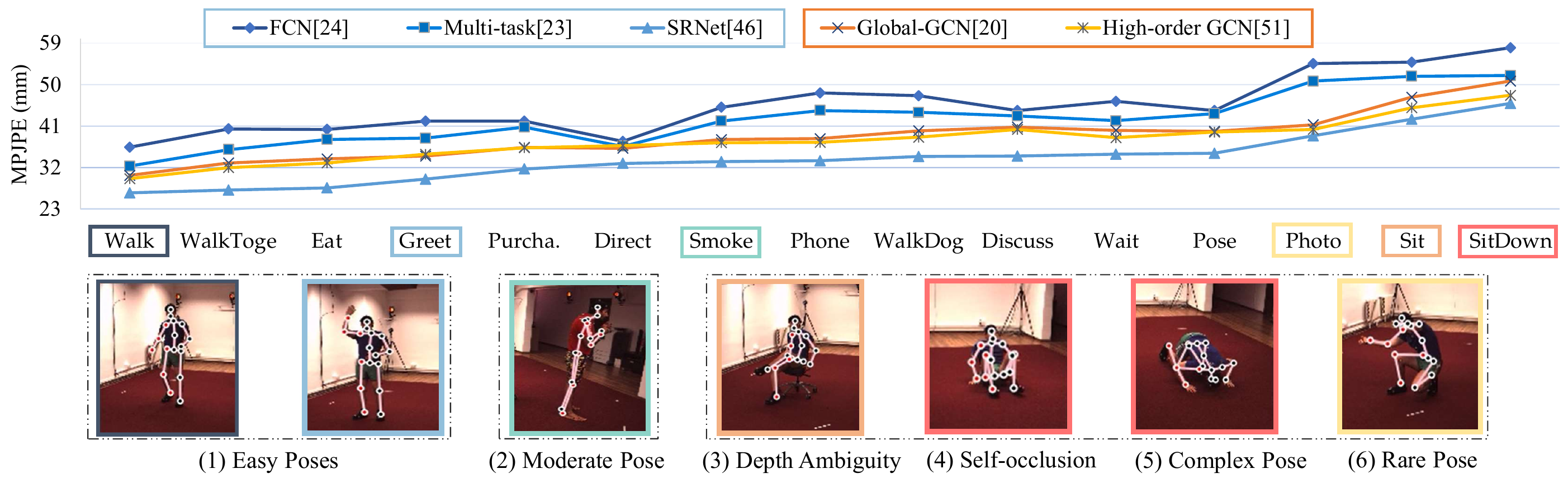}
\end{center}
\vspace{-0.5cm}
\caption{The examples of two \emph{easy poses}, one \emph{moderate pose}, and four kinds of \emph{hard poses} in 2D-to-3D pose estimation with 2D detected poses as inputs (shown in the images). Although rapid progress has been made in this field, both \textcolor[RGB]{133,181,213}{non-graph} methods and \textcolor[RGB]{231,110,47}{graph-based} ones yield large prediction error on these \emph{hard poses}.}
\vspace{-0.5cm}
\label{fig:ob1}
\end{figure*}

Single-view skeleton-based 3D pose estimation problem plays an important role in numerous applications, such as human-computer interaction, video understanding, and human behavior analysis.
Given the 2D skeletal positions detected by a 2D keypoint detector (e.g.,~\cite{chen2018cascaded,newell2016stacked,sun2019deep}), this task aims to regress the 3D positions of the corresponding joints. It is a challenging task and has drawn lots of attention from academia in recent years.  Even though the average performance increases steadily over the years, the prediction errors on some poses are still quite high. 

Fig.~\ref{fig:ob1} shows some examples in a widely used dataset, Human3.6M~\cite{ionescu2013human3}.
Some actions (e.g., ``Sit" and ``Sit Down") contain many poses with depth ambiguity, self-occlusion, or complex poses. Also, 
there inevitably exist some poses rarely seen in the training dataset. Similar to the definition of hard examples in object detection~\cite{lin2017focal} and semantic segmentation~\cite{li2017not}, we collectively regard those poses with high prediction errors as \emph{hard poses}. 

Early attempts~\cite{martinez2017simple,pavllo20193d} simply use fully-connected networks (FCN) to lift the 2D keypoints into 3D space. However, the dense connection of FCN is prone to overfit, leading to relatively poor performance. To tackle this problem, geometric dependencies are incorporated into the network in~\cite{fang2018learning,park20183d,wang2019generalizing,zeng2020srnet}, which significantly improve prediction accuracy. As articulated human body can be naturally modeled as a graph, with the recent development of graph neural networks (GNN)~\cite{kipf2016semi,li2019deepgcns,abu2019mixhop,xu2018representation,zhang2020hop}, various GNN-based methods~\cite{zhao2019semantic,ci2019optimizing,liu2020learning,cai2019exploiting,zou2020high} are proposed in the literature for 2D-to-3D pose estimation.

GNN-based solutions naturally capture the relationship between body joints. For a target node, aggregating features from its neighboring nodes facilitates bringing in semantic information to relieve the uncertainty in estimating its 3D position. In other words, for the estimation of a particular node in the graph (e.g., left hand), both its direct neighbor (i.e., left elbow) and other nodes that are multiple hops away in the graph (e.g., left shoulder and even right foot in some poses) may 
provide useful information that contributes to the position estimation of the target node, and the learning 
of skeletal graph neural networks is to capture such context information for better 2D-to-3D pose estimation.  However, existing GNN-based solutions do not fully tap the potential of the skeleton graph. The reasons are two-fold:
\vspace{-0.2cm}
\begin{itemize}
    \item The power of graph neural networks lies in the aggregation of neighboring nodes, which, however, contributes both useful information and undesired noises. On the one hand, aggregating distant nodes in the skeletal graph does provide useful information; On the other hand, the more distant the nodes, the more likely undesired noises are introduced into the aggregation procedure. Existing works do not consider such signal-to-noise issues in message passing over the GNN.
    \vspace{-0.2cm}
    \item The relationship between body joints varies with different poses. For example, for poses in ``running'', the hand-foot joints are closely related, while for poses in ``Sitting", there is no such strong relationship. It is rather difficult to capture such information with a static skeleton graph across all poses. 
\end{itemize}
\vspace{-0.2cm}

This work proposes novel skeletal GNN learning solutions to mitigate the above problems, especially for hard poses. Our contributions are summarized as follows:
\vspace{-0.2cm}
\begin{itemize}

    \item We propose a \emph{hop-aware} hierarchical \emph{channel-squeezing} fusion layer to extract relevant information from neighboring nodes effectively while suppressing undesired noises. This is inspired by the feature squeezing works~\cite{zhu2019rethinking,yu2019autoslim,zeng2021hop}, wherein channel size is reduced to keep valuable information in each layer. Specifically, we squeeze long-range context features (i.e., information from distant nodes) and fuse them with short-range features in a hierarchical manner. 

    \vspace{-0.2cm}

    \item Inspired by GNN-based action recognition work~\cite{shi2019skeleton,ye2020dynamic,liu2020disentangling}, we build dynamic skeletal graphs, wherein the edges between nodes are not only from the fixed human skeleton topology but also the node features to capture action-specific poses. To cope with the change of dynamic graphs over time and relieve outliers from frame-level features, we further integrate temporal cues into the learning process of dynamic graphs. The proposed temporal-aware dynamic graph construction procedure is robust and effective for 2D-to-3D pose estimation.  
    \vspace{-0.2cm}
\end{itemize}

We conduct experiments on Human3.6M dataset~\cite{ionescu2013human3}, and the proposed solution outperforms state-of-the-art techniques~\cite{zeng2020srnet} in 3D pose estimation by 10.3\% on average, and greatly improves on \emph{hard poses}. Compared to state-of-the-art GNN-based solutions, we surpass ~\cite{ci2019optimizing} by \textbf{16.3}\%. As the proposed method is a plug-and-play module, we further integrate it into the skeleton-based action recognition framework, achieving state-of-the-art performance.

\section{Preliminaries and Motivation}

This work focuses on GNN-based 3D pose estimation. We first describe the general skeletal GNN construction procedure in Sec.~\ref{sec:relate_sg}. Next, we discuss existing GNN-based solutions for 3D pose estimation in Sec.~\ref{sec:relate_pose}. Finally, Sec.~\ref{sec:obs} motivates this work.

\begin{figure}
\begin{center}
\includegraphics[width=0.45\textwidth]{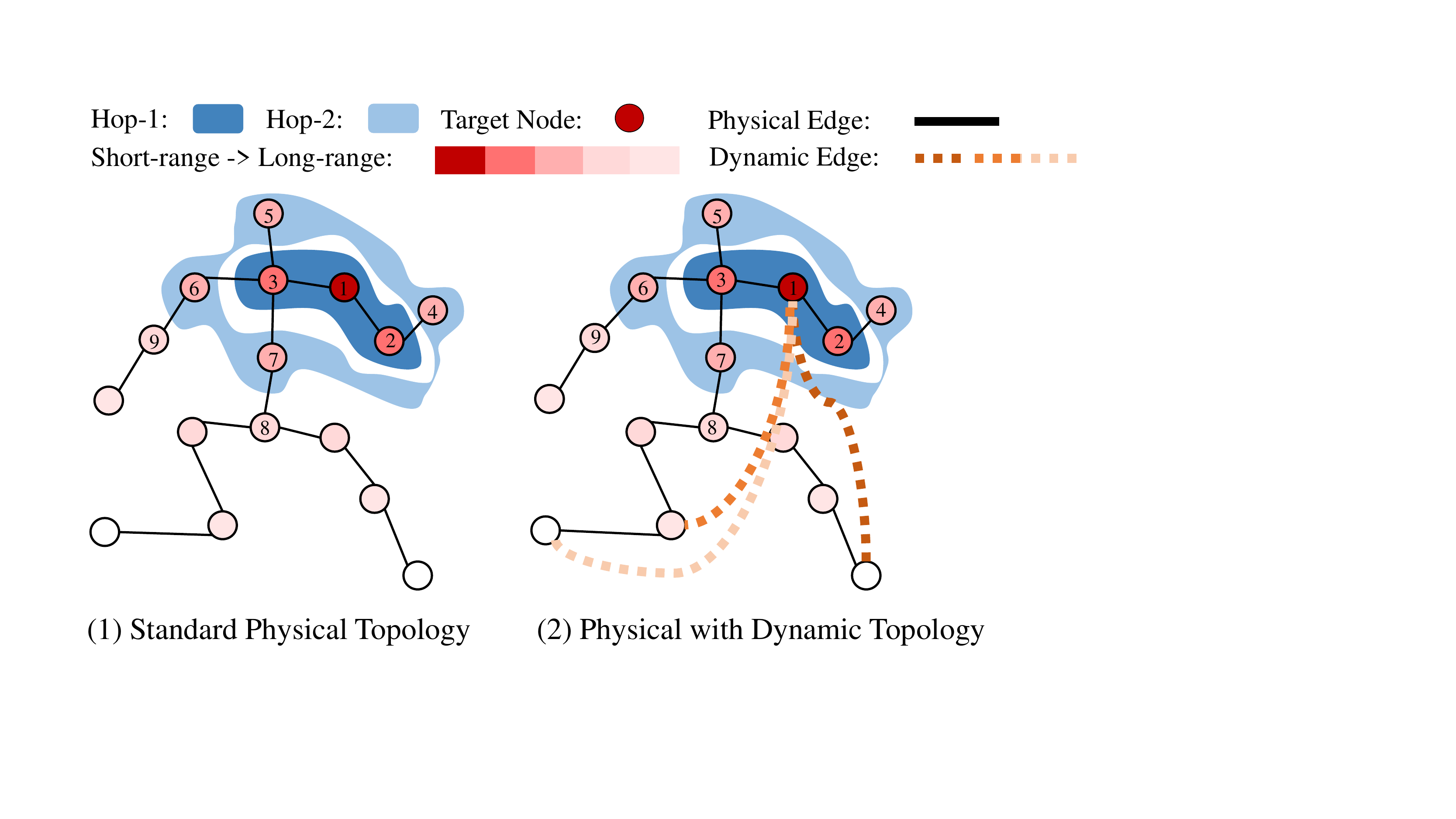}
\end{center}
\vspace{-0.5cm}
\caption{The illustration of a skeletal graph with human physical edges and action-specific dynamic edges.}
\vspace{-0.5cm}
\label{fig:hop}
\end{figure}

\subsection{Skeletal Graph Neural Network}
~\label{sec:relate_sg}
The human skeleton can be naturally modeled as a graph as shown in Fig.~\ref{fig:hop}(a). The nodes in the graph represent the 2D positions of human joints, and the edges between two joints denote bone connections. Hop-$k$ demonstrates the shortest path $k$ between two nodes. For instance, take node 1 (right shoulder) as the target node, node $2$ (right elbow) and node $3$ (neck) are its hop-$1$ neighbors (masked in dark blue); while nodes $4$ (right hand), $5$ (head), $6$ (left shoulder), and $7$ (spine) are its hop-$2$ neighbors (masked in light blue).

In a graph neural network, 
adjacency matrix determines the information passing among nodes, and the objective of the learning procedure is to obtain each node's features by aggregating features from its neighboring nodes. As hop-$k$ increases during message passing, the information passed from the corresponding neighbors varies from short-range context to long-range context. 

\subsection{GNN-Based 3D Human Pose Estimation}
\label{sec:relate_pose}

Recently, various GNN-based methods~\cite{zhao2019semantic,ci2019optimizing,cai2019exploiting,liu2020comprehensive,liu2020learning,zou2020high} are proposed in the literature for 3d pose estimation. As weight sharing strategy restricts the representation power, a locally connected network (LCN)~\cite{ci2019optimizing} is proposed to learn the weight of each node individually for enhancing the representation differences among nodes, achieving better generalization capability.

High-order GCN~\cite{zou2020high} explores different aggregation methods on high-order neighbors to capture the long-range dependencies among nodes. 
However, it may introduce more noises from less-related nodes without differentiating the impacts between short-range and long-range contexts.

Furthermore, some recent works~\cite{zhao2019semantic,ci2019optimizing} try to learn the edge weights of the skeletal graph. However, without changing the graph topology, the effectiveness of such a dynamic solution is limited, especially for rare poses.

\subsection{Observation and Motivation}
~\label{sec:obs}
Fig.~\ref{fig:ob1} shows the prediction accuracy of recent 3D pose estimation methods~\cite{martinez2017simple,pham2019unified,zeng2020srnet,liu2020learning,zou2020high}. As can be observed from this figure, most of them suffer from poor performance for some complex actions, such as, ``Sit," ``Sit Down," and ``Take Photos."
We attribute this phenomenon to that these hard poses require both short-range and long-range context information for better estimation. Meanwhile, existing solutions do not fuse them effectively in the learning process. 

\textbf{Distant neighbors pass not only valuable semantic information but also irrelevant noise.}
Existing GNN-based methods try to aggregate the semantic information in both short-range and long-range neighbors. However, they ignore that the messages passed from distant neighbors also contain irrelevant noise. For example, "Walking" has a clear pattern, which contains strong correlation between arm and leg. Intuitively, we need to take both of them into account. However, due to some personalized style, besides the action pattern, there is also some heterogeneous noise. Therefore, it is necessary to suppress such irrelevant noise. Interestingly, through experiments, we observe that such noise is sensitive to the channel dimension. The channel dimension can constrain the amount of information and noise passed among nodes in a skeleton graph. In other words, an effective channel squeezing strategy could filter out undesired noise while keeping valuable information. Consequently, we propose a hop-aware hierarchical channel-squeezing transform on long-range features to improve aggregation effectiveness in skeletal GNN learning. 

\textbf{Dynamic graph construction is useful but should be delicately designed.}~Existing GNN-based methods construct the graph based on the physical skeleton topology~\cite{ci2019optimizing,zhao2019semantic}, like Fig.~\ref{fig:hop}(a). However, the strong hidden relationships among nodes vary with actions. By constructing a dynamic graph as shown in Fig.~\ref{fig:hop}(b), such a relationship might be more useful than physical features. Although the dynamic graphs seem intuitive for representing different motion-specific relations, it usually seems vulnerable to the single-frame outliers. Thus, we introduce temporal information to make dynamic graph learning robust.

\section{Method}

\begin{figure}[t]
\begin{center}
\vspace{-0.4cm}

\includegraphics[width=0.47\textwidth]{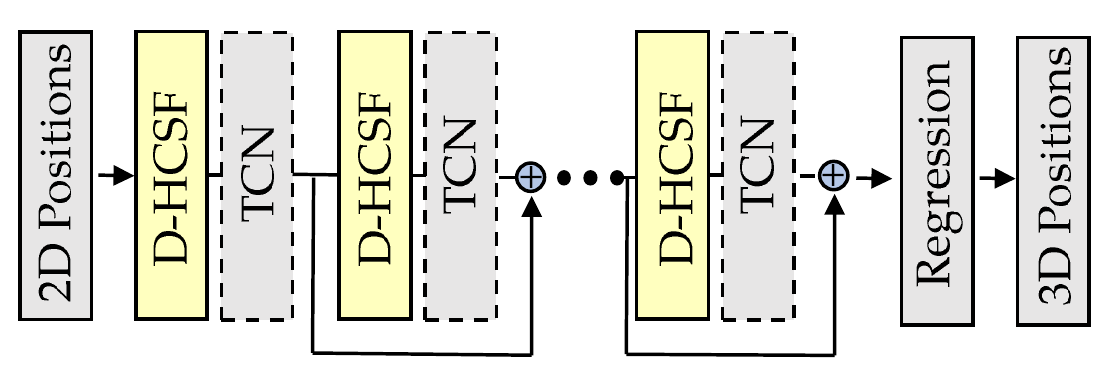}
\end{center}
\vspace{-0.5cm}
\caption{The framework of our method. The key is a specially designed module called Dynamic Hierarchical Channel-Squeezing Fusion layer (\emph{D-HCSF}), shown in Fig.~\ref{fig:branch} with details.}
\vspace{-0.5cm}

\label{fig:frame}
\end{figure}

\begin{figure}[t]
\begin{center}

\scriptsize
\includegraphics[width=0.43\textwidth]{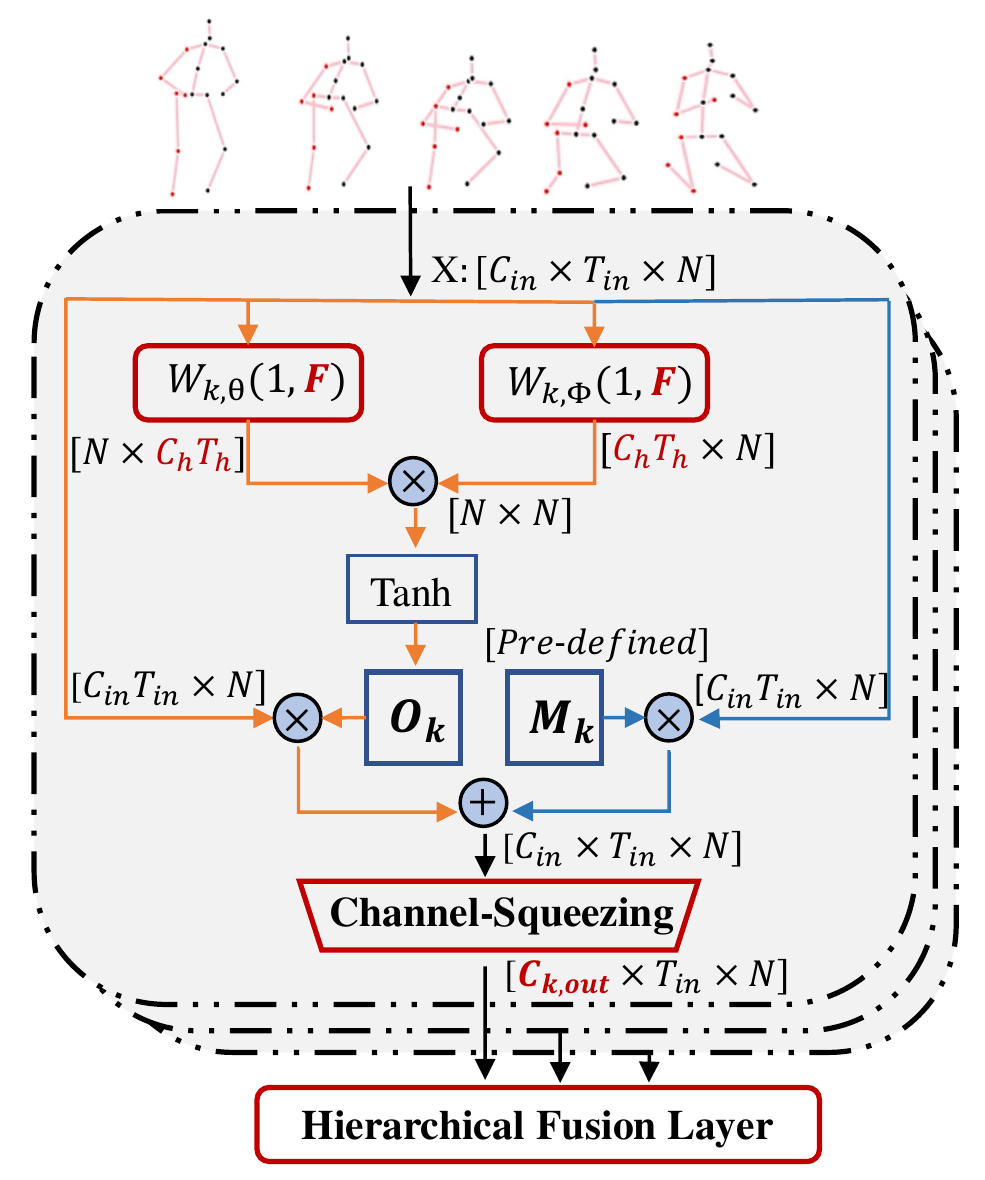}
\end{center}
\vspace{-0.5cm}
\caption{The architecture of our Dynamic Hierarchical Channel-Squeezing Fusion (D-HCSF) layer under $k$ hops. Each dotted box consists of two streams: a weighted graph learning branch based on fixed physical edges (blue lines) and a dynamic graph learning branch to update the graph based on the node features adaptively (orange lines).}
\vspace{-0.5cm}
\label{fig:branch}
\end{figure}

In this work, our goal is to reduce errors of 3D human pose estimation, especially on \emph{hard poses}. More specifically, given 2D keypoints $X\in \mathbb{R}^{N \times 2}$, with $N$ nodes, the model outputs better 3D positions $Y\in \mathbb{R}^{N \times 3}$.
The framework is designed based on the observations and motivations in Sec.~\ref{sec:obs}, and shown in Fig~\ref{fig:frame}. The core of our framework is the module: a Dynamic Hierarchical Channel-Squeezing Fusion Layer (\emph{D-HCSF}) shown in Fig~\ref{fig:branch}. It contains a hierarchical channel-squeezing fusion scheme for updating features of each node (Fig~\ref{fig:branch}), and a temporal-aware dynamic graph learning component for updating the dynamic graph.

In this section, we first revisit the formulation of generic GCN~\cite{kipf2016semi} and LCN~\cite{ci2019optimizing} in Sec.~\ref{sec:gnn_ske}. Then, we introduce our hierarchical channel-squeezing fusion scheme in Sec.~\ref{sec:llcn}. Finally, we propose the dynamic graph learning and consider temporal-aware strategy in this process in Sec~\ref{sec:ldcn}.

\subsection{Vanilla Graph Neural Network}
\label{sec:gnn_ske}
Given a graph $\mathcal{G}=(\mathcal{V},\mathcal{E})$, it consists of the nodes $\mathcal{V}$ and the edges $\mathcal{E}$. We revisit a generic GCN~\cite{kipf2016semi} layer defined as follows:
\begin{equation}
    \mathbf{H} = \sigma(\Hat{\mathbf{A}} \mathbf{X} \mathbf{W}),
    \label{eq:mp}
\end{equation}
where $\mathbf{A}\in \mathbb{R}^{N \times N}$ is an adjacency matrix with $N$ nodes, indicating the connections between nodes. If the joint $j$ is dependent on the joint $i$, then ${a_{ij}=1}$. Otherwise, the connections are set to zero ${a_{ij}=0}$. We denote the input node features as $\mathbf{X} \in \mathbb{R}^{N \times C_{in}}$, the learnable weight matrix as $\mathbf{W} \in \mathbb{R}^{C_{in} \times C_{out}}$, and the activation function as $\sigma(\cdot)$. \emph{For simplification, we ignore the $\sigma(\cdot)$ in the following formulas}.

The GCN's representation power is limited by weight sharing strategy in node regression problem, while Ci et al.~\cite{ci2019optimizing} propose a locally connected network (LCN), which introduces the node-wise trainable parameters to enhance the differences among node features.~This aggregation scheme learns different relations among different nodes. Accordingly, we recap its basic formulation. For clarity purposes, we take the embedding learning of node $i$ from the direct neighbors in a layer as an example:

\begin{equation}
\mathbf{h}_i = \sum_{j \in \mathcal{N}_{1,i}} (\Hat{a}_{ij} \mathbf{x}_j \mathbf{W}_{ij}),
\label{eq:lcn2}
\end{equation}

where $\mathcal{N}_{1,i}$ contains the self-node and direct neighbors (hop=1) of the node $i$. We denote $\Hat{a}_{ij}$ as the value of $i_{th}$ row and $j_{th}$ column in the adjacency matrix $\Hat{\mathbf{A}}$, which distinctly aggregates features among neighbors. $\mathbf{x}_j$ is the inputs of the neighbor $j$. $\mathbf{W}_{ij} \in \mathbb{R}^{C_{in} \times C_{out}}$ denotes the learnable weights between the node pair $(i,j)$, and $\mathbf{h}_i$ is the updated features of the node $i$. Hence, the final output $\mH$ is represented by the concatenation of all node features.

\begin{figure*}[t]
\begin{center}
\includegraphics[width=0.9\textwidth]{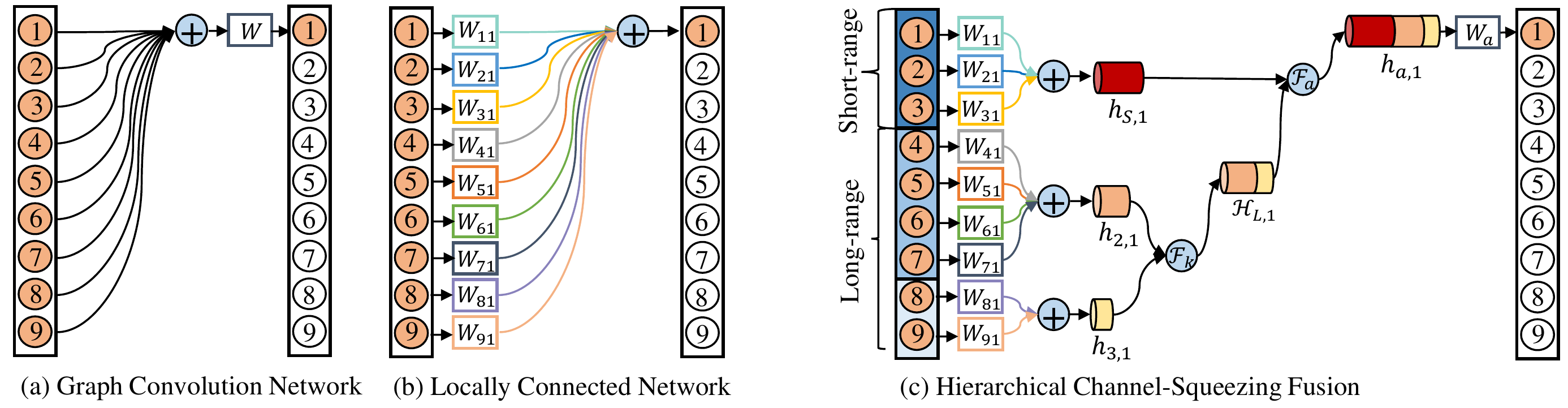}
\end{center}
\vspace{-0.5cm}
\caption{The architectures of (a) Graph Convolution Network (GCN)~\cite{kipf2016semi}, (b) Locally Connected Network (LCN)~\cite{ci2019optimizing}, and (c) Our Hierarchical Channel-Squeezing Fusion (HCSF). We take the feature updating of node 1 as an example. The index of the node corresponds to Fig.~\ref{fig:hop}.}
\vspace{-0.5cm}

\label{fig:fuse}
\end{figure*}

\subsection{Hierarchical Channel-Squeezing Fusion Layer}
\label{sec:llcn}
Inspired by the first observation in Sec.~\ref{sec:obs}, we find (i) hierarchical spatial features are important to capture better short-to-long range context; (ii) distinguishing short-range and long-range context in fusion strategies is necessary to relieve irrelevant long-range context while keeping their essential components in hard pose estimation. Thus, as illustrated in Fig.~\ref{fig:fuse} (c), we propose a hierarchical channel-squeezing fusion layer to reach the above hypothesis. 

\vspace{-0.4cm}
\paragraph{\textbf{Hierarchical fusion layer.}}
\label{sec:hflayer}
~Accordingly, we take node-wise aggregation LCN~\cite{ci2019optimizing} as a baseline.~To capture different ranges in spatial context, we generalize Eq.~\ref{eq:lcn2} by modifying the direct neighbors $\mathcal{N}_{1,i}$ to hop-$k$ neighbors $\mathcal{N}_{k,i}$. Then, we can get updated features $\mathbf{h}_{k,i}$ from hop-$k$ neighbors.

To integrate multi-hop features in a layer, we propose a hierarchical fusion block as follows. It consists of two parts. First, we consider the short-range features $\vh_{S,i}$ within hop-$S$, which contain the most essential context of the target node $i$. We thus keep the whole information without squeezing them. We then define a farthest hop $L$ to obtain the potentially useful information as a set of long-range context $\cH_{L,i}$ defined as: 
\vspace{-0.2cm}
\begin{equation}
\mathbf{h}_{S,i} = \sum_{ j \in \mathcal{N}_{S,i}}(  \Hat{a}_{ij} \mathbf{x}_j \mathbf{W}_{ij}),\\
\label{eq:hopS}
\end{equation}
\vspace{-0.2cm}
\begin{equation}
\cH_{L,i} = \{\vh_{k,i} | k=S+1,...,L\},
\label{eq:hopL}
\end{equation}
where $S$ is less than or equal to $L$, set empirically.

To fuse features from different contexts, we introduce two fusion functions, namely $\cF_k$ and $\cF_a$, to form a two-stage procedures. $\cF_k$ first transforms a set of long-range features $\cH_{L,i}$ to obtain a fused long-range features $\vh_{L, i}$, and then $\cF_a$ fuses $\vh_{L, i}$ with short-range features $\vh_{S, i}$ to get output $\vh_{a,i}$. We refer to such two-step fusion scheme as \emph{hierarchical fusion block}. Finally, we process the feature $\vh_{a,i}$ through a transformation $\mathbf{W}_a$ to obtain an final output $\mathbf{h}_i$ with pre-defined dimension. Formally, the final output $\mathbf{h}_i$ of this fusion layer is defined as:

\begin{equation}
\mathbf{h}_i  = \mathcal{F}_a [\mathbf{h}_{S,i},  \mathcal{F}_k(\cH_{L,i} )]\mathbf{W}_a.
\label{eq:fuse_hop}
\end{equation}

\vspace{-0.3cm}

\paragraph{\textbf{Channel-Squeezing block.}}
\label{sec:csblock}
To retain useful information while suppressing the irrelevant information of long-range context, we hypothesize that the context contains less relevant information. Hence, we propose a set of bottleneck transformations, named Channel-Squeezing Blocks to filter irrelevance by an end-to-end learning scheme.

To reach the above hypothesis, we distinguish the learnable matrix $\mathbf{W}_{ij} \in \mathbb{R}^{C_{in} \times C_{k,out}}$ from its output size $C_{k,out}$ of long-range context $\cH_{L,j}$.
Moreover, based on the theoretical Information Gain (IG) analysis on the hop-$k$~\cite{gao2021} measured by the average KL-divergence, where the information gain IG($k$) of hop $k$ always fits well on the function $ae^{-bk}$, where $a,b > 0$. It means the information gain from long-range context will be decreased exponentially. In our condition, we propose a simplified relation function to decide the output size $C_{k,out}$ for each long-range context to reflect the decay of useful information as $k$ increases. 
\vspace{-0.2cm}
\begin{equation}
    C_{k,out} = d ^ {(k-L)} * C_{in},
\label{eq:decay}
\end{equation}
where $d \in [0 , 1]$ indicates the channel squeezing ratio, and $k \in [S+1, L]$. 
For short-range context $\vh_{S,i}$, the output channels of $\mW_{i,j}$ keep the same as $C_{in}$ without squeezing to reduce the loss of useful features. Compared with Eq.~\ref{eq:lcn2}, LCN is a special kind of HCSF module when $C_{k,out}$ is a constant across hops, and $\mathcal{F}_k, \mathcal{F}_a$ are summations.

Please noted that the HCSF layer could also adapt to other graph-based frameworks, like GCN~\cite{kipf2016semi}, Pre-Agg, Post-Agg~\cite{liu2020learning}, the difference lies in weight sharing schemes, which is orthogonal to our approach.

\subsection{Temporal-aware Dynamic Graph Learning}
\label{sec:ldcn}

In this subsection, we present a framework when dealing with dynamic skeleton topology. Based on the second observation in Sec.~\ref{sec:obs}, with a static graph, the model is hard to properly reflect the relations, especially for \emph{hard poses}.

\vspace{-0.5cm}
\paragraph{\textbf{Learning dynamic graph.}}
We learn the dynamic graph from two streams, where one is to update by physical skeleton topology, and the other is to update through features of nodes. For the first stream (Fig.~\ref{fig:branch} blue lines), we initialize the graph $\mathbf{M}_k$ as the physical skeleton topology following SemGCN~\cite{zhao2019semantic}. Then, this graph will be updated during training. After training, this graph $\mathbf{M}_k$ keeps fixed during inference and reflects the distribution of the whole dataset.  

As the poses change, the connections should be varied during testing to capture motion-specific relations. Hence, we introduce a scheme for the second stream (Fig.~\ref{fig:branch} orange lines) to learn the graph from input features dynamically. Then, the graph can be adapted to the input pose during the inference. The input feature $\mathbf{X}$ is separately transformed by $\mathbf{W}_{k, \theta}$ and $\mathbf{W}_{k, \phi}$ for each branch. After multiplying two transformed features, we obtain a $N\times N$ matrix indicating relations among nodes.

\vspace{-0.3cm}
\begin{equation}
    \mathbf{O}_k = \mathcal{F}\{[(\mathbf{X})^{tr} \mathbf{W}_{k, \theta}] [(\mathbf{W}_{k, \phi})^{tr} \mathbf{X}]\},
    \label{eq:ms_soft}
\end{equation}
where $\mathcal{F}$ denotes the activation function (i.e., Tanh), and $tr$ denotes the transpose of a matrix. The output $\mathbf{O}_k$ generates unique connections changing adaptively with input $\mathbf{X}$. 

Since learning dynamic graphs directly from input features may be sensitive to outliers (e.g., jitter and missing joints). We regard $\mathbf{O}_k$ as dynamic offsets to refine the weighted graph $\mathbf{M}_k$. 
Hence, the final formulation of the dynamic graph of Hop-$k$ is as follows:
\vspace{-0.2cm}
\begin{equation}
    \mathbf{A}_{k} = \mathbf{M}_k+ \alpha\mathbf{O}_k,
    \label{eq:dy2}
\end{equation}
where $\alpha $ is a learnable scalar to adjust the scale of dynamic offsets. Then, we can aggregate the hop-$k$ features by $\mathbf{A}_{k}$, and the following fusion block is the same as Eq.~\ref{eq:fuse_hop}.

\vspace{-0.5cm}
\paragraph{\textbf{Temporal-aware dynamic graph.}}
\label{sec:temp_gcn}
The connections between nodes may naturally vary over space and time, and the frame-level dynamic graph will suffer from unstable inputs, it is potential to optimize the dynamic graph learning process combining spatial and temporal context as a whole. In terms of introducing temporal information, the input will add a time dimension, where $\mathbf{X}\in \mathbb{R}^{C_{in} \times T_{in} \times N}$.

We develop a temporal-aware scheme in Eq.~\ref{eq:ms_soft}. Instead of only applying linear transformations in spatial domain, we integrate temporal context by 1D temporal convolution layer with a kernel size of $1\times F$. It can smooth and filter the spatial outliers in input $\mathbf{X}$ to make the learning process robust.

\section{Experiment}

In this section, we perform experimental studies to demonstrate the effectiveness of the proposed solution. 
\vspace{-4pt}

\subsection{3D Human Pose Estimation}
\label{sec:3dpose}

\subsubsection{Data Description}
\vspace{-0.2cm}

\noindent -- \emph{Human3.6M}~\cite{ionescu2013human3} consists of 3.6 million video frames with $15$ actions from $4$ camera viewpoints, where accurate 3D human joint positions are captured from high-speed motion capture system. Following previous works~\cite{ci2019optimizing,cai2019exploiting,zeng2020srnet,martinez2017simple}, we adopt the standard cross-subject protocol with 5 subjects (S1, S5, S6, S7, S8) as training set and another 2 subjects (S9, S11) as test set. It is commonly evaluated by two metrics, namely \emph{the mean per joint position error (MPJPE)} with 17 joints of each subject and \emph{the Procrustes Analysis MPJPE (PA-MPJPE)} to relieve the inherent scale, rotation, and translation problems.

\noindent -- \emph{MPI-INF-3DHP} \cite{mehta2017monocular,mehta2017vnect} contains both constrained indoor scenes and complex outdoor scenes, covering a greater diversity of poses and actions, where it is usually taken as a cross-dataset setting to verify the generalization ability of the proposed methods. For evaluation, we follow common practice~\cite{ci2019optimizing,wang2019generalizing,zeng2020srnet} by using the Percentage of Correct Keypoints (PCK) with a threshold of 150mm and the Area Under Curve (AUC) for a range of PCK thresholds. 
\vspace{-0.2cm}

\begin{table*}[t]
\begin{center}
\scriptsize

\resizebox{\textwidth}{!}
{
\begin{tabular}{ l c c c c c c c c c c c c c c c c }

\hline
Method & Direct & Discuss & Eat & Greet & Phone& Photo & Pose  & Purcha. & Sit & SitD & Smoke & Wait & WalkD & Walk & WalkT & Avg.\\
\hline
Luvizon et al. \cite{luvizon20182d}& 63.8 & 64.0 & 56.9 & 64.8 & 62.1 & 70.4 & 59.8 & 60.1&71.6 & 91.7 &60.9 & 65.1 & 51.3 & 63.2 & 55.4 & 64.1  \\ 
Martinez et al. \cite{martinez2017simple}& 51.8 &56.2 &58.1 &59.0 &69.5 &78.4 &55.2 &58.1& 74.0& 94.6 &62.3 &59.1& 65.1& 49.5& 52.4 &62.9 \\
Park et al.\cite{park20183d}&49.4 &54.3& 51.6& 55.0& 61.0 &73.3 &53.7 &50.0& 68.5& 88.7 &58.6 &56.8& 57.8& 46.2 &48.6 &58.6\\
Wang et al. \cite{wang2019generalizing}&47.4 &56.4 &49.4 &55.7& 58.0& 67.3& 46.0& 46.0& 67.7& 102.4& 57.0& 57.3& 41.1& 61.4& 40.7 &58.0\\
Zhao et al. \cite{zhao2019semantic}$\dagger$&47.3 &60.7& 51.4& 60.5& 61.1 &49.9 &47.3 &68.1& 86.2& 55.0& 67.8& 61.0& 42.1& 60.6& 45.3& 57.6\\
Zou et al. \cite{zou2020high}$\dagger$&49.0& 54.5& 52.3 &53.6 &59.2& 71.6& 49.6& 49.8& 66.0& 75.5& 55.1& 53.8& 58.5& 40.9& 45.4& 55.6\\
Liu et al. \cite{liu2020learning}$\dagger$&48.4& 53.6& 49.6& 53.6 &57.3 &70.6 &51.8 &50.7& 62.8 &74.1 &54.1 &52.6 &58.2& 41.5& 45.0& 54.9\\
Ci et al. \cite{ci2019optimizing}$\dagger$&46.8 &52.3& 44.7& 50.4& 52.9& 68.9& 49.6& 46.4& 60.2 &78.9& 51.2& 50.0& 54.8& 40.4& 43.3& 52.7\\
Liu et al. \cite{liu2020comprehensive}$\dagger$&46.3& 52.2& 47.3& 50.7& 55.5 &67.1 &49.2 &46.0 &60.4 &71.1& 51.5 &50.1 &54.5 &40.3 &43.7& 52.4\\
Pavllo et al. \cite{pavllo20193d} &47.1& 50.6& 49.0& 51.8 &53.6 &61.4& 49.4 &47.4 &59.3 &67.4 &52.4& 49.5& 55.3& 39.5& 42.7& 51.8\\
Cai et al. \cite{cai2019exploiting}$\dagger$&46.5 &48.8 &47.6& 50.9& 52.9 &61.3 &48.3 &45.8 &59.2 &64.4& 51.2& 48.4& 53.5& 39.2& 41.2& 50.6\\
Zeng et al. \cite{zeng2020srnet}&44.5& \textbf{48.2} &47.1 &47.8 &51.2 &\textbf{56.8} &50.1& \textbf{45.6}& 59.9 &66.4 &52.1 &45.3 &54.2 &39.1 &40.3 &49.9\\
\hline
\emph{Ours-HCSF} $\dagger$&43.4&49.7&45.1&47.6&50.7&57.5&47.1&45.9&56.5&61.1&49.8&\textbf{47.1}&\textbf{51.4}&35.8&37.8&	48.4\\
\emph{Ours-HCSF w/A}$\dagger$ &\textbf{43.1}&50.4&\textbf{43.9}&\textbf{45.3}&\textbf{46.1}&57.0&\textbf{46.3}&47.6&\textbf{56.3}&\textbf{61.5}&\textbf{47.7}&47.4&	53.5&\textbf{35.4}&\textbf{37.3}&\textbf{47.9}\\
\hline\hline
Martinez et al. \cite{martinez2017simple}&37.7 &44.4 &40.3 &42.1& 48.2& 54.9 &44.4 &42.1 &54.6& 58.0 &45.1 &46.4 &47.6 &36.4 &40.4& 45.5\\
Pham et al. \cite{pham2019unified} &36.6 &43.2 &38.1 &40.8 &44.4 &51.8 &43.7 &38.4 &50.8 &52.0 &42.1 &42.2 &44.0 &32.3 &35.9 &42.4\\
Zhao et al. \cite{zhao2019semantic}$\dagger$ &37.8 &49.4 &37.6 &40.9 &45.1 &41.4 &40.1& 48.3& 50.1 &42.2& 53.5 &44.3 &40.5 &47.3& 39.0 &43.8\\
Wang et al. \cite{wang2019generalizing}& 35.6 &41.3 &39.4 &40.0 &44.2 &51.7 &39.8 &40.2 &50.9 &55.4 &43.1 &42.9 &45.1 &33.1 &37.8 &42.0\\
Liu et al. \cite{liu2020learning}$\dagger$&36.2 &40.8 &33.9 &36.4 &38.3 &47.3 &39.9 &34.5 &41.3 &50.8 &38.1 &40.1 &40.0& 30.3& 33.0 &38.7\\
Liu et al. \cite{liu2020comprehensive}$\dagger$&36.8 &40.3& 33.0 &36.3 &37.5 &45.0 &39.7 &34.9 &40.3 &47.7& 37.4& 38.5& 38.6& 29.6 &32.0 &37.8\\
Ci et al. \cite{ci2019optimizing} $\dagger$&36.3& 38.8 &29.7& 37.8& 34.6& 42.5& 39.8& 32.5&  36.2&  39.5& 34.4& 38.4& 38.2& 31.3& 34.2&36.3\\
Zeng et al. \cite{zeng2020srnet}& 32.9 & 34.5 & 27.6 & 31.7 & 33.5 &42.5 & 35.1 & 29.5& 38.9& 45.9 & 33.3 & 34.9 & 34.4 & 26.5 & 27.1& 33.9\\
\hline
\emph{Ours-HCSF} $\dagger$&29.0&34.1&27.3&31.7&\textbf{28.8}&\textbf{34.8}&34.4&\textbf{27.3}&33.5&38.9&\textbf{30.4}&32.3&\textbf{29.7}&24.6&25.2&30.8\\
\emph{Ours-HCSF w/A} $\dagger$&\textbf{26.8}&\textbf{33.2}&\textbf{26.7}&\textbf{30.0}&30.8&36.7&\textbf{31.5}&27.4&\textbf{33.1}&\textbf{38.0}&30.8&\textbf{31.8}&30.3&\textbf{23.9}&\textbf{25.0}&\textbf{30.4}\\
\hline
\end{tabular}}
\end{center}
\vspace{-0.5cm}
\caption{Comparison of single-frame 3D pose estimation in terms of MPJPE on Human3.6M. Works above the double line show results from detected 2D poses, and the below results are from 2D ground truth inputs to explore the upper bound of these methods. We highlight the graph-based methods by $\dagger$. w/A denotes using dynamic graphs. Best results in bold.}
\vspace{-0.5cm}
\label{tab:human_p1_t}
\end{table*}

\subsubsection{Method Comparison}
\vspace{-0.1cm}
In terms of the frames of inputs, \emph{single-view} 3D pose estimation can be divided into \emph{single-frame} and \emph{temporal} settings. We first compare our \textbf{HCSF} module and \textbf{Dynamic graph learning} with other previous works under the \emph{single-frame} setting. Then, we extend to the \emph{temporal} setting to compare related works with our \textbf{Temporal-aware dynamic graph learning with HCSF} scheme.

\vspace{-0.4cm}
\paragraph{\textbf{Comparison with \emph{Single-frame} methods.}}
As shown in Tab.~\ref{tab:human_p1_t}, we compare our methods with other baselines. 
Under the standard protocol with \emph{2D detected inputs}~\cite{chen2018cascaded}, our methods can improve the graph-based method~\cite{cai2019exploiting} from 50.6mm to 47.9mm (relative 5.3\% improvement), and surpass the non-graph based method~\cite{zeng2020srnet} by 2.0mm (relative 4.0\% improvement). Since the results of 2D detected poses would affect uncertainty, it is better to consider using 2D ground truth as input to explore the upper bound of these methods. Accordingly, with \emph{2D ground truth inputs}, our proposed model improves the graph-based state-of-the-art~\cite{ci2019optimizing} from 36.3mm to 30.4mm (relative \textbf{16.3}\% improvement). Although LCN~\cite{ci2019optimizing} aggregates long-range ($L=3$) information to relieve depth ambiguities, it ignores the fact that distant joints may bring more disruptions while they still contain certain useful information. The proposed HCSF module considers this effect by squeezing the different hop features into different latent spaces and then hierarchically fusing them. Moreover, our method surpasses state-of-the-art non-GNN method~\cite{zeng2020srnet} by 3.5mm (relative \textbf{10.3}\% improvement), which further proves the effectiveness among the general methods.

\vspace{-0.5cm}
\paragraph{\textbf{Comparison with \emph{Temporal} methods.}}
We compare with temporal methods with nine frames as inputs and Tab.~\ref{tab:temporal} shows the comparison in terms of average error.
For all methods, we select their reported results with similar input frames for comparison.
The result shows that the proposed method can outperform previous approaches consistently.
With temporal-aware dynamic graph construction, the proposed solution further improves the result by 0.7mm.

\begin{table*}[htbp]
\begin{center}
\small
\resizebox{\textwidth}{!}
{
\begin{tabular}{ l| c |c |c |c |c |c|c }
\hline
Method &Hossain et al.~\cite{rayat2018exploiting} & Lee et al. \cite{lee2018propagating} & Pavllo et al.~\cite{pavllo20193d} & Cai et al.~\cite{cai2019exploiting} & Lin et al.~\cite{lin2019trajectory} & Ours w/o \emph{T} & Ours \\
\hline
MPJPE (mm) &58.3&52.8&49.8&48.8&48.8&46.4&\textbf{45.7}\\
\hline
\end{tabular}}
\end{center}
\vspace{-0.5cm}
\caption{Comparison on \emph{temporal 2D detected poses input} with similar input frames ($5, 3, 9, 7, 10, 9, 9$ frames, individually) for comparison. The noted w/o \emph{T} denotes using dynamic graphs without temporal-aware scheme.}
\vspace{-0.5cm}
\label{tab:temporal}
\end{table*}

\vspace{-0.5cm}
\paragraph{\textbf{Improvements on hard poses.}}

As discussed earlier, we define hard poses as those with high prediction errors and \emph{they are model-specific}. That is, while hard poses have some inherent characteristics (e.g., depth ambiguity and self-occlusion), they are handled differently with different models~\cite{ci2019optimizing,zhao2019semantic,zeng2020srnet,zou2020high}. Consequently, one pose that exhibits large errors on one model may show satisfactory results on another model and vice versa (we show more detailed analyses in the supplemental material). However, statistically speaking, if a model handles hard poses better, it would have the following properties: (1) those actions with high prediction errors would be improved more; (2) the proportion of poses with high errors would be smaller; (3) the upper bound of high-error poses would be smaller. 

\begin{figure}[htbp]
\begin{center}
\includegraphics[width=0.47\textwidth]{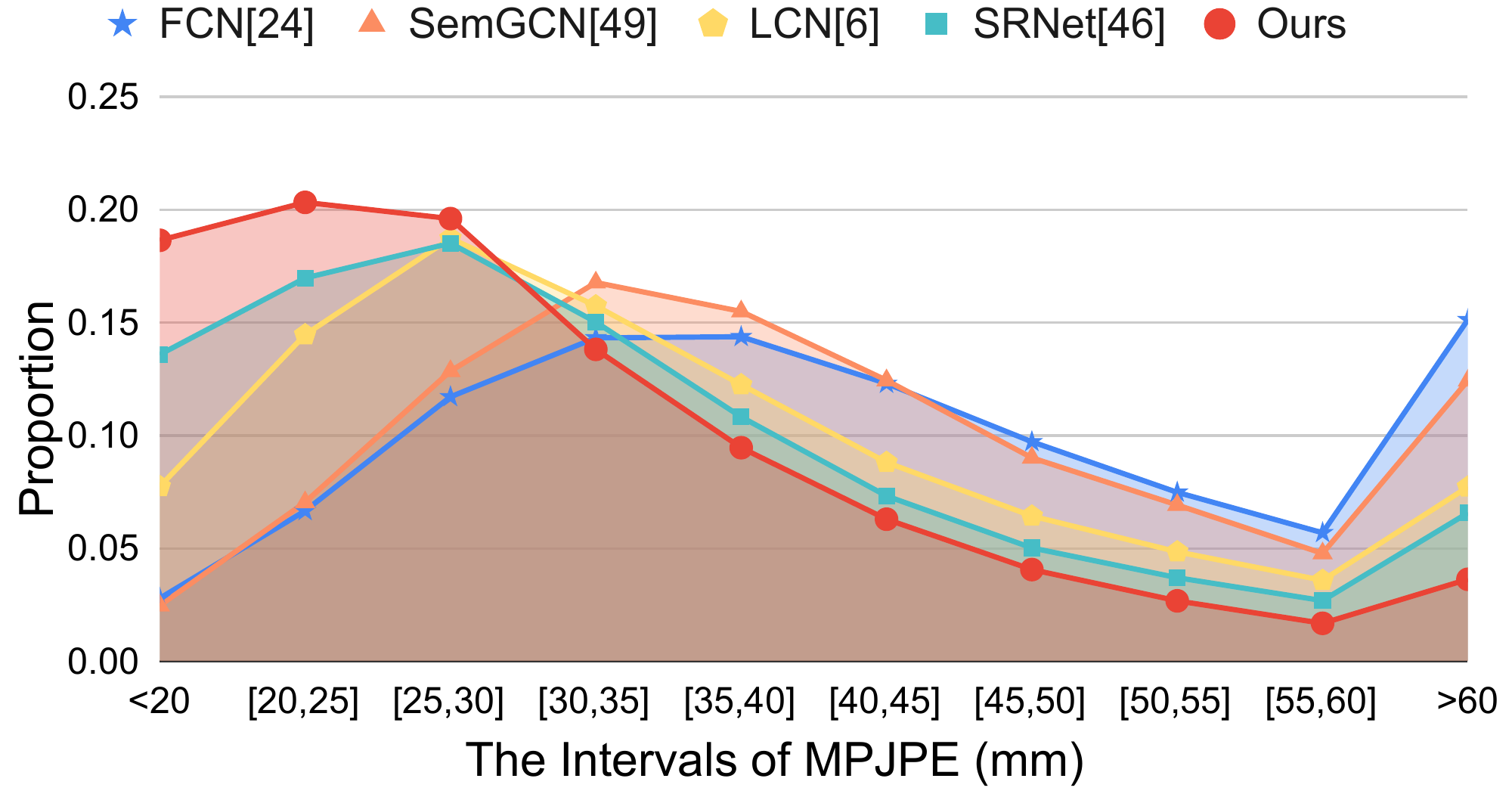}
\end{center}
\vspace{-0.6cm}
\caption{MPJPE distribution on the testset of Human3.6M. }
\vspace{-0.4cm}
\label{fig:mp}
\end{figure}

Compared with state-of-the-art solution~\cite{zeng2020srnet}, our method reduces the prediction errors by 7.9mm, 6.1mm, 5.8mm, and 5.8mm (relative 17.2\%, 18.5\%, 14.9\%, 13.6\% improvements) on the actions ``SitDown", ``Direct", ``Sit" and ``Photo", respectively. The average improvement of the hard poses is \textbf{16.1}\% in Tab.~\ref{tab:human_p1_t}. Next, in Fig.~\ref{fig:mp}, we compare the error distribution in the test set with four existing solutions~\cite{martinez2017simple,zhao2019semantic, ci2019optimizing, zeng2020srnet}. We can observe that there are much fewer poses with high prediction errors with our proposed solution. Specially, there are only 3.6\% cases with MPJPE above 60mm with our solution, while it is more than 6\% with all the other methods. In fact, the number of cases with MPJPE above 40mm is consistently lower, and the number of cases with MPJPE less than 30mm is consistently higher with our solution than that with other methods.

\vspace{-0.2cm}

\begin{figure}[htbp]
\begin{center}
\vspace{-0.4cm}
\includegraphics[width=0.4\textwidth]{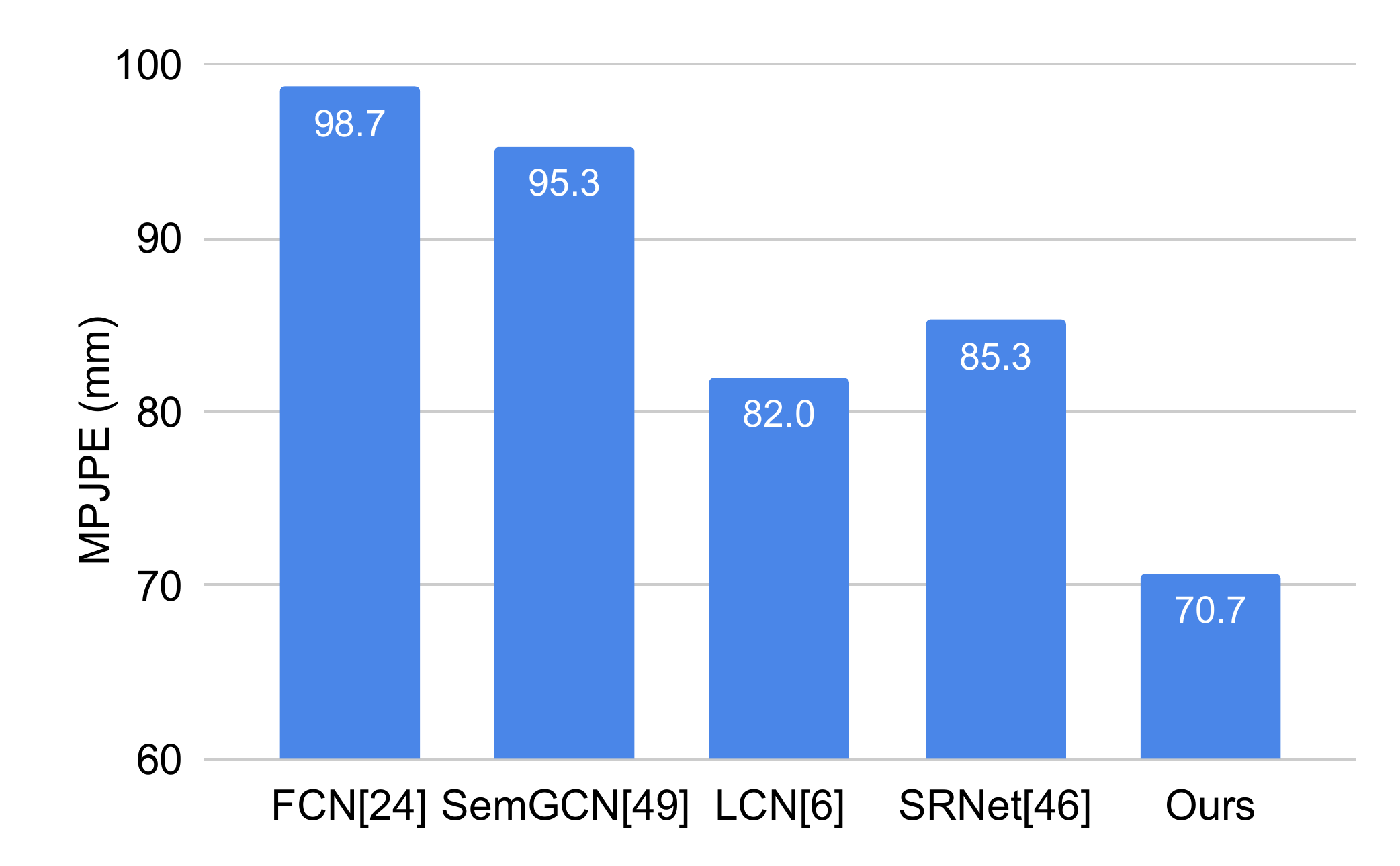}
\end{center}
\vspace{-0.7cm}
\caption{Mean-Error comparison of the 5\% Hardest Poses.}
\vspace{-0.3cm}
\label{fig:upp}
\end{figure}

Last, we present the mean errors for the top $5$\% hardest poses of five methods in Fig.~\ref{fig:upp}, ours is $\textbf{70.7}mm$, which is \textbf{$13.8\%$} and \textbf{$17.1\%$} smaller than the SOTA methods LCN ($82.0mm$) and SRNet ($85.3mm$), respectively. 

\begin{figure}[htbp]
\begin{center}
\includegraphics[width=0.47\textwidth]{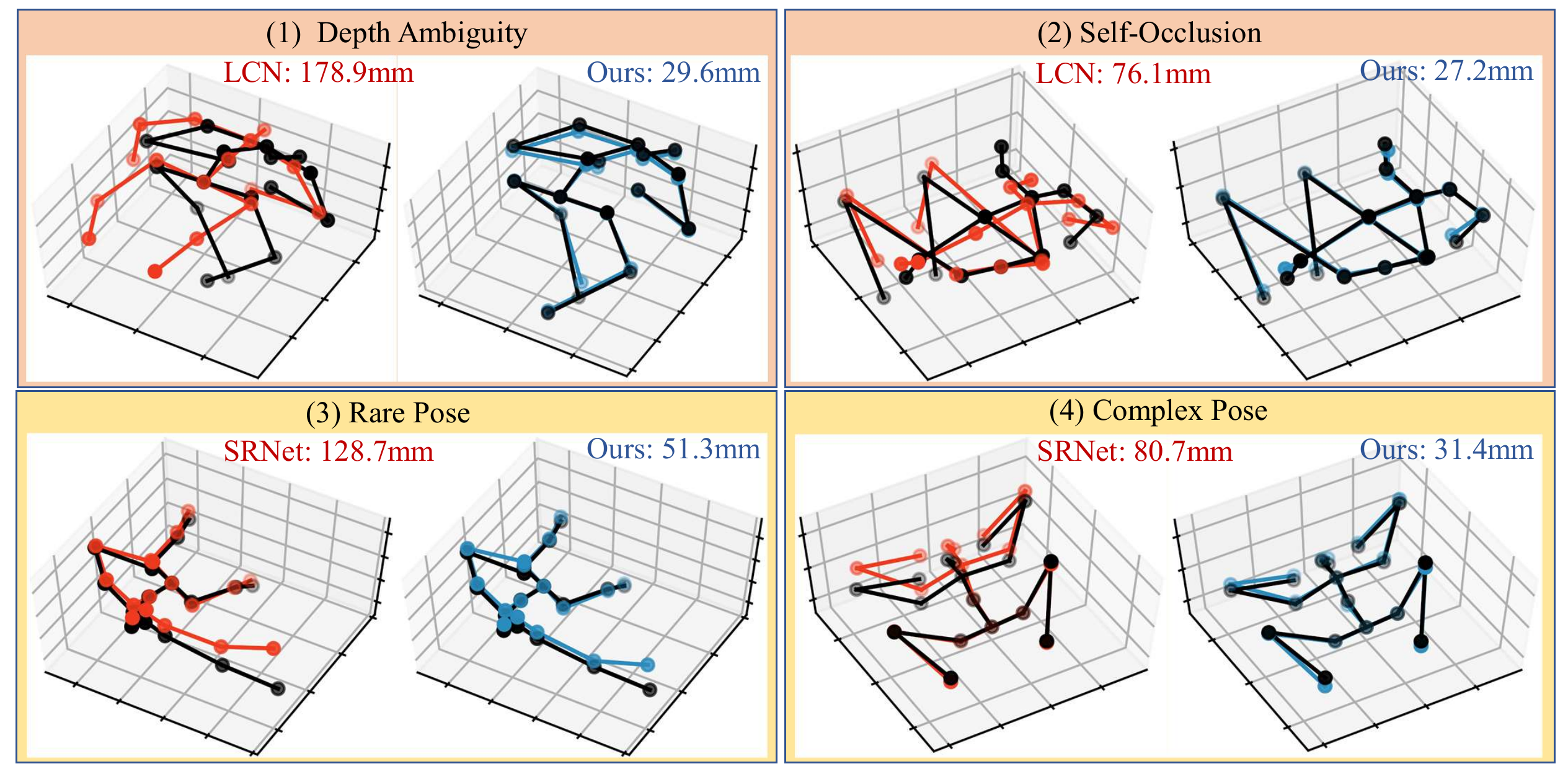}
\vspace{-0.5cm}
\end{center}
\caption{Qualitative results of hard poses.~3D ground truth, \textcolor[RGB]{ 192,0,0}{SOTA methods}, and \textcolor[RGB]{47,85,151}{ours} are black, \textcolor[RGB]{ 192,0,0}{red}, \textcolor[RGB]{47,85,151}{blue} in order.}
\vspace{-0.3cm}
\label{fig:viz}
\end{figure}

We also show visualization results in Fig.~\ref{fig:viz}, compared with SOTA methods (upper LCN, below SRNet). In summary, the above results demonstrate the benefits of the proposed technique on hard poses.

\vspace{-0.5cm}
\paragraph{\textbf{Comparison on MPI-INF-3DHP.}}~We further test our model trained with the Human3.6M dataset on the MPI-INF-3DHP dataset to verify its generalization capability. Tab.~\ref{tab:com} shows about $5.5\%$ improvements on all metrics over related methods. 

\begin{table}[htbp]
{
\scriptsize{
\begin{center}
\scalebox{0.95}
{
\begin{tabular}{ l |c| c|c|c|c}
\hline
Method &FCN~\cite{martinez2017simple}&OriNet~\cite{luo2018orinet}&LCN~\cite{ci2019optimizing}&SRNet~\cite{zeng2020srnet}&Ours\\
\hline
Outdoor &31.2&65.7&77.3&80.3&$\textbf{84.6}\uparrow_{5.4\%}$\\
\hline
All PCK &42.5&65.6&74.0&77.6&$\textbf{82.1}\uparrow_{5.8\%}$\\
\hline
All AUC& 17.0&33.2&36.7&43.8&$\textbf{46.2}\uparrow_{5.5\%}$\\
\hline
\end{tabular}}
\end{center}}}
\vspace{-0.5cm}
\caption{Cross-dataset results on MPI-INF-3DHP.}
\vspace{-0.5cm}
\label{tab:com}
\end{table}

\subsection{Ablation Study}
To study some important design choices in the proposed method, we take 2D ground truth poses as inputs and adopt MPJPE as an evaluation metric for analysis.

\vspace{-0.5cm}
\paragraph{\textbf{Impact of hierarchical fusion.}}
Tab.~\ref{tab:dis} shows that (1) two hierarchical design consistently outperforms \emph{Non-hierarchy} by 2mm$\sim$3.7mm under different $L$, indicating that it is not appropriate to fuse long-range and short-range information in a single stage;
(2) without considering different contributions from different hops, the performance of \emph{Hierarchy w/o hop-aware} is inferior to \emph{Hierarchy}, leading to a consistent performance drop of 1.6mm$\sim$1.8mm. It illustrates that the importance of processing long-range contexts according to hops. We fix $S$=1, $d$=$1$/$16$ by default.

\begin{table}
\begin{center}
\scalebox{0.85}
{
\begin{tabular}{ l| c | c |c |c |c }
\hline
$L$ &2& 3 & 4 & 5 & 6\\
\hline
\emph{Non-hierarchy} &37.2&36.2&37.7&37.4&39.9\\
\emph{Hierarchy w/o hop-aware} &34.6&34.2&34.7&35.4&36.2\\
\emph{Hierarchy}&-&\textbf{32.6}&\textbf{32.9}&\textbf{33.7}&\textbf{34.4} \\
\hline

\end{tabular}}
\end{center}
\vspace{-0.5cm}
\caption{Comparison of aggregating without hierarchy strategy (\emph{Non-hierarchy}), with hierarchy scheme but regarding all hops information equally (\emph{Hierarchy w/o hop-aware}) and our full design (\emph{Hierarchy}). }
\vspace{-0.5cm}
\label{tab:dis}
\end{table}

\vspace{-0.5cm}
\paragraph{\textbf{Impact of squeezing ratio $d$.}}

We set $S=1, L=2$ to study the influence of the squeezing ratio $d$.
Tab.~\ref{tab:low_dim_lcn} shows that as $d$ decreases, the corresponding MPJPE first decreases and then increases. As $d$ controls the output channel size of different hop features, the small value of $d$ indicates the small output dimension for channel-squeezing transform. The decrease of $d$ may first reduce the irrelevant information from long-range context, and thus reducing the pose estimation error. When $d$ takes an extreme value $\frac{1}{16}$, the useful information may also be substantially squeezed, leading to a performance drop.

\begin{table*}[htbp]
\vspace{-0.5cm}
\begin{center}
\small
\resizebox{\textwidth}{!}
{
\begin{tabular}{ l| c |c |c |c |c|c|c|c}
\hline
Method & Baseline & a & b & c & d & e & f & g\\
\hline
$\mA_k$ & $ori$ &Only $\mM_k$ ($ori$) & Only $\mM_k$ ($dense$)&Only $\mM_k$ ($rand$) & Only $\mO_k$ & $\mM_k + \mO_k$ & Eq.~\ref{eq:dy2} & Eq.~\ref{eq:dy2} w/T\\
\hline\hline
MPJPE(mm)&30.8&32.1&35.7&40.8&44.3&30.5&30.4&\textbf{29.7}\\
\hline

\end{tabular}}
\end{center}
\vspace{-0.5cm}
\caption{Comparison on the effects of dynamic graph learning $\mA$. $ori$ is the static graph with physical connections. \emph{Baseline} takes $\mA_k$ as $ori$. \emph{Only $\mM_k$ ($\cdot$)} denotes applying $\mM_k$ with different initialization. \emph{Only $\mO_k$} keeps the dynamic offset in Eq.~\ref{eq:dy2}. \emph{$\mM_k + \mO_k$} equals to set $\alpha=1$ in Eq.~\ref{eq:dy2}. \emph{w/T} represents the temporal-aware scheme defined in Sec.~\ref{sec:ldcn}.}
\label{tab:dyn}
\vspace{-0.5cm}
\end{table*}  

\begin{table}
\begin{center}
\scalebox{0.9}
{
\begin{tabular}{ l |c|c|c|c|c}
\hline

$d$ &$1$&$1/2$&$1/4$&$1/8$& $1/16$\\
\hline
MPJPE (mm) & 35.4&34.7&33.8&\textbf{31.4}&32.6\\
\hline
\end{tabular}}
\end{center}
\vspace{-0.5cm}
\caption{Influence of the squeezing ratio $d$.}
\vspace{-0.6cm}
\label{tab:low_dim_lcn}
\end{table}

\begin{table}[hb]
\begin{center}
\vspace{-0.5cm}
\scalebox{0.9}
{
\begin{tabular}{ l| c |c |c |c |c|c}
\hline
$L$ & 1 & 2 &3&4 & 5&6\\
\hline
$S=0$& \textbf{35.5}&34.6&34.8&34.7&35.4&36.1\\

$S=1$& -&\textbf{32.6}&\textbf{33.4}&\textbf{34.0}&\textbf{34.7}&\textbf{35.7}\\
$S=2$& -&-&34.9&35.1&35.6&36.2\\
\hline
LCN \cite{ci2019optimizing} &38.0&35.7&36.2&37.7&37.4&39.9\\
\hline
\end{tabular}}
\end{center}
\vspace{-0.5cm}
\caption{The influence of the hop-$S$ and the hop-$L$.}
\vspace{-0.5cm}
\label{tab:kl}
\end{table}

\vspace{-0.5cm}
\paragraph{\textbf{Impact of $S$ and $L$.}}
\label{sec:sr-lcn}

As shown in Tab.~\ref{tab:kl}, $S=1$ yields consistently good results under different $L$. As features within hop-$S$ will not be squeezed, it is in line with our intuition that the direct neighbors provide the most relevant information to the target node.
Besides, a random combination of $S$ and $L$ surpasses the strong baseline LCN~\cite{ci2019optimizing}, demonstrating the effectiveness of our design.

\vspace{-0.5cm}
\paragraph{\textbf{Impact of the dynamic graph learning}}
In Sec.~\ref{sec:ldcn}, we introduce a dynamic graph $\mA_k$ consisting of a learnable graph $\mM_k$ and a graph offset $\mO_k$ learned from input features.
As Tab.~\ref{tab:dyn}a, ~\ref{tab:dyn}b and ~\ref{tab:dyn}c illustrate, the initialization of the dynamic graph learning is important. The result shows that it is hard for all connected initialization (Tab.~\ref{tab:dyn}b) and random (Tab.~\ref{tab:dyn}c) to converge well. On the other hand, taking the physical topology as an initial graph (Tab.~\ref{tab:dyn}a) can achieve better results ($32.1$mm).
Only learning the dynamic offsets (Tab.~\ref{tab:dyn}d) leads to severe performance drop. Relying only on input features may weaken its capability of dealing with data noise.

Tab.~\ref{tab:dyn}e adds the weighted graph $\mM_k$ and dynamic offsets $\mO_k$, obtaining a 0.3mm performance gain over \emph{baseline}.
Moreover, although the dynamic graph shows priority in representing different motion-specific relations, it is usually vulnerable to the single-frame outliers. After considering the temporal context, we can further improve the baseline from 30.8mm to 29.7mm.

\subsection{Skeleton-based Action Recognition}
\vspace{-0.1cm}

As a plug-and-play module, we integrate the proposed HCSF layer and temporal-aware dynamic graph learning into \emph{skeleton-based action recognition}.
Given a sequence of human skeleton coordinates in videos, this task categorizes them into a predefined action class.
We conduct experiments on two commonly used datasets, namely, \textbf{NTU RGB+D 60}~\cite{shahroudy2016ntu} and \textbf{NTU RGB+D 120}~\cite{liu2019ntu}. Due to the limits of contents, we leave detailed datasets, implementation details, and ablation study in the supplemental material.

Following related works~\cite{yan2018spatial,shi2019skeleton,shi2020skeleton,liu2020disentangling,cheng2020skeleton,ye2020dynamic}, we employ a GNN model with ten spatial-temporal convolution neural layers as the baseline framework.~We adopt the proposed framework shown in Fig.~\ref{fig:frame} to this task by changing the final regression branch into classification branch. Our results use the multi-stream framework as in~\cite{shi2020skeleton,cheng2020skeleton}. Single stream results are \emph{in the supplemental material.}

Tab.~\ref{tab:ntu} shows that our solution outperforms SOTA on both benchmarks. In particular, the top-1 accuracy is $89.2\%$ and $87.5\%$ in the X-Set and X-Sub settings on the more complex dataset~\cite{liu2019ntu}, surpassing state-of-the-art solutions irrespective of the fact that they employ sophisticated attention modules~\cite{shi2020skeleton} and temporal-dilated combinations in each layer~\cite{liu2020disentangling}. Our method can construct robust dynamic relations through the proposed HCSF layer and temporal-aware scheme. Therefore, those inherent relations among joints are better captured, enhancing the capability to distinguish different actions. 

\begin{table}[htbp]
\begin{center}
\vspace{-0.3cm}
\scriptsize
{
\begin{tabular}{c|cc|cc}
\hline
\multirow{2}{*}{Method} & \multicolumn{2}{c|}{\textbf{NTU RGB+D 60}}& \multicolumn{2}{c}{\textbf{NTU RGB+D 120}}\\  
& X-Sub (\%)  & X-View (\%) &   X-Sub (\%)  &  X-Set (\%)  \\
\hline
ST-GCN~\cite{yan2018spatial} &84.3&92.7&71.3&72.4\\
AS-GCN~\cite{li2019actional}&86.8&94.2&77.7&78.9\\
SGN~\cite{zhang2020semantics}&89.0&94.5&79.2&81.5\\
2s-AGCN~\cite{shi2019skeleton}&88.9&95.1&82.9&84.9\\
NAS-GCN~\cite{peng2020learning}&89.4&95.7&-&-\\
Mix-dim~\cite{peng2020mix}&89.7&96.0&80.5&83.2\\
ST-Transformer~\cite{plizzari2020spatial}&89.3&96.1&82.7&84.7\\
MS-AAGCN~\cite{shi2020skeleton}&90.0&96.2&-&-\\
Shift-GCN~\cite{cheng2020skeleton}&90.7&96.5&85.9&87.6\\
MMDGCN~\cite{xia2021multi}&90.8&96.5&86.8&88.0\\
DecoupleGCN~\cite{cheng2020eccv}&90.8& 96.6&86.5&88.1\\
MS-G3D~\cite{liu2020disentangling}& 91.5&96.2& 86.9& 88.4\\
\hline
Ours&\textbf{91.6}&\textbf{96.7}&\textbf{87.5}&\textbf{89.2}\\
\hline
\end{tabular}}
\end{center}
\vspace{-0.5cm}
\caption{Comparison against state-of-the-art methods on the NTU RGB+D 60 and 120 Skeleton dataset in terms of Top-1 accuracy(\%). Best results in bold.}
\vspace{-0.6cm}
\label{tab:ntu}
\end{table}

\section{Conclusion}
\vspace{-0.15cm}
In this work, we explore a novel skeleton-based representation learning method for better dealing with hard poses in 3D pose estimation. The proposed Hierarchical Channel-Squeezing Fusion module can encode short-range and long-range contexts, keeping essential information while reducing irrelevant noises. Besides, we learn dynamic graphs that adaptively evolved based on the input poses rather than relying on a fixed pre-defined graph. 
Our method surpasses all non-graph methods by $10.3\%$ and enhances the effectiveness of graph-based methods. We hope that our method would inspire the field of skeleton-based representation learning.

\vspace{-0.15cm}
\section*{Acknowledgements}
\vspace{-0.15cm}
This work was supported in part by the Innovation and Technology Fund (ITF) under Grant No. MRP/022/20X, Innovation and Technology Commission, Hong Kong S.A.R.

{\small
\bibliographystyle{ieee_fullname}
}

\pagebreak

\twocolumn[{
	\renewcommand\twocolumn[1][]{#1}
	\begin{center}
		\textbf{\Large Supplementary Material:\\Learning Skeletal Graph Neural Networks for Hard 3D Pose Estimation}\end{center}
}]
\vspace{10pt}
\setcounter{equation}{0}
\setcounter{figure}{0}
\setcounter{table}{0}
\setcounter{section}{0}
\setcounter{page}{0}
This supplementary material presents more experimental details, including data pre-processing, implementation details, additional experimental results, and ablation studies.

\section{3D Human Pose Estimation}
\vspace{-0.2cm}
In this section, we demonstrate more detailed results on 3D human pose estimation. Sec.~\ref{sec:hm_imp} gives more details on experiment settings. Second, Sec.~\ref{sec:hard_model} analyzes features of hard poses in this task. Third, Sec.~\ref{sec:hm_pa} compares existing methods by the metric of PA-MPJPE. 
Finally, Sec.~\ref{sec:hm_ab} shows the ablation study of only using a dynamic graph with HCSF module.
\subsection{Dataset and Implementation Details} 
\label{sec:hm_imp}
\subsubsection{Dataset Pre-processing} 
\vspace{-0.2cm}
We follow our baseline~\cite{ci2019optimizing} to transform the 3D joint position under the camera coordinate system into the pixel coordinate system to remove the influence of pose scales for the single-view pose estimation. Following previous works~\cite{pavllo20193d,ci2019optimizing,zeng2020srnet}, we normalize 2D input poses in the range of [-1, 1] according to the width and height of images. The furthest hop is $6$ in our pre-defined topology. Meanwhile, we set the entry values of the adjacency matrix to be ones if two nodes are physically connected and zero if not.
\vspace{-0.2cm}
\subsubsection{Training Details} 
\vspace{-0.2cm}
We build a six-layer network as the basic setting, including the first layer, two cascaded blocks, and the last layer. For a single-frame setting, each cascaded block consists of two HCSF layers followed by BN, LeakyReLU (alpha is $0.2$), and dropout (random drop probability is $0.25$). Besides, each block is wrapped with a residual connection, as shown in Fig.$3$ in the main paper. The channel size of each layer we report in the final result is $128$. In the ablation study, we set all output channels as $64$ for each node. The above framework is a common structure that is also used in those works~\cite{martinez2017simple,pavllo20193d,ci2019optimizing,zhao2019semantic,zeng2020srnet}. For temporal settings, each cascaded block consists of one HCSF layer and one TCN layer. The fusion functions $\cF_k$ and $\cF_a$ are concatenation operators by default, which can also be addition, multiplication. L1 regression loss is used between the ground truth and outputs. Moreover, we train our model for $80$ epochs using Adam~\cite{kingma2014adam} optimizer. The initial learning rate is set as $0.001$, and the exponential decay rate is $0.95$. The mini-batch size is $256$. For data augmentation, we follow~\cite{pavllo20193d,ci2019optimizing,zhao2019semantic,zeng2020srnet} and use horizontal flip data augmentation at both training and test stages. Then, we evaluate our method with standard protocol following ~\cite{ci2019optimizing,zhao2019semantic,zeng2020srnet,pavllo20193d}.

\subsection{Further Analysis on Model-Specific Hard Poses}
\label{sec:hard_model}
\begin{figure}[thbp]	
	\subfigure[The hard poses of FCN~\cite{martinez2017simple}] 
	{
		\begin{minipage}{8cm}
			\centering          
			\includegraphics[width=\linewidth]{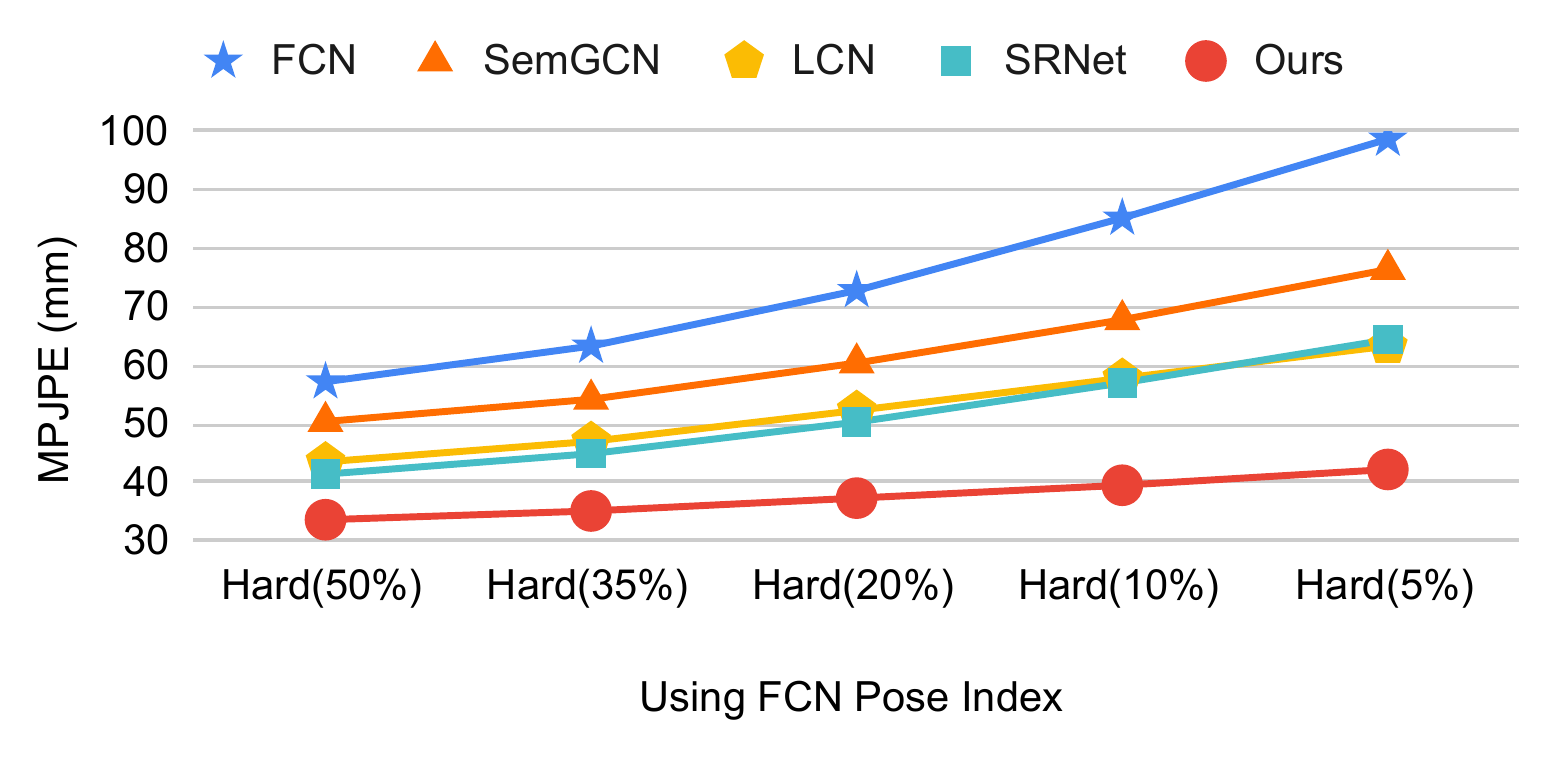}   
		\end{minipage}
	}
	\vspace{-0.3cm}
	\subfigure[The hard poses of SemGCN~\cite{zhao2019semantic}] 
	{
		\begin{minipage}{8cm}
			\centering      
			\includegraphics[width=\linewidth]{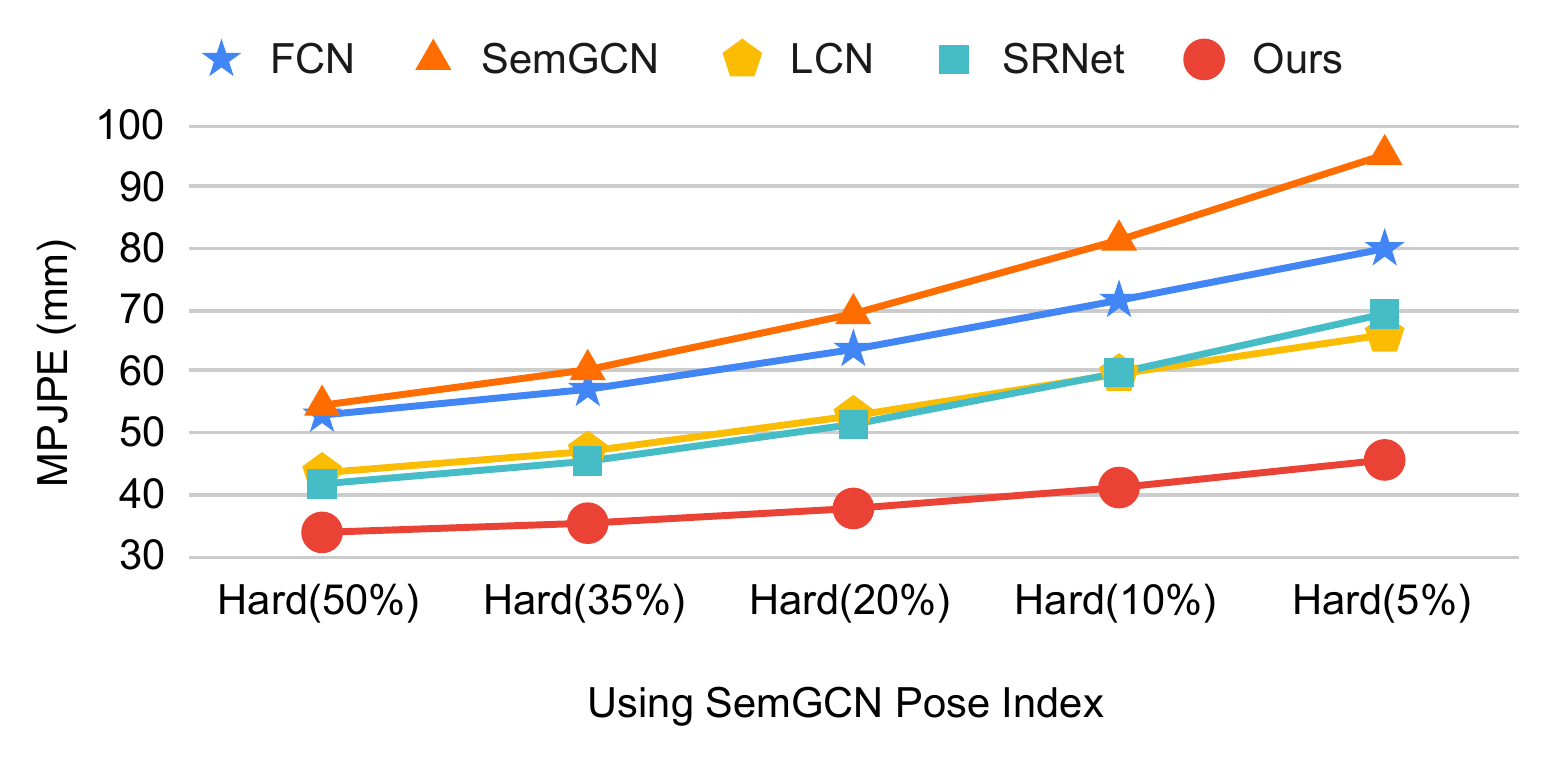}   
		\end{minipage}
	}
	\vspace{-0.3cm}
	\subfigure[The hard poses of LCN~\cite{ci2019optimizing}] 
	{
		\begin{minipage}{8cm}
			\centering      
			\includegraphics[width=\linewidth]{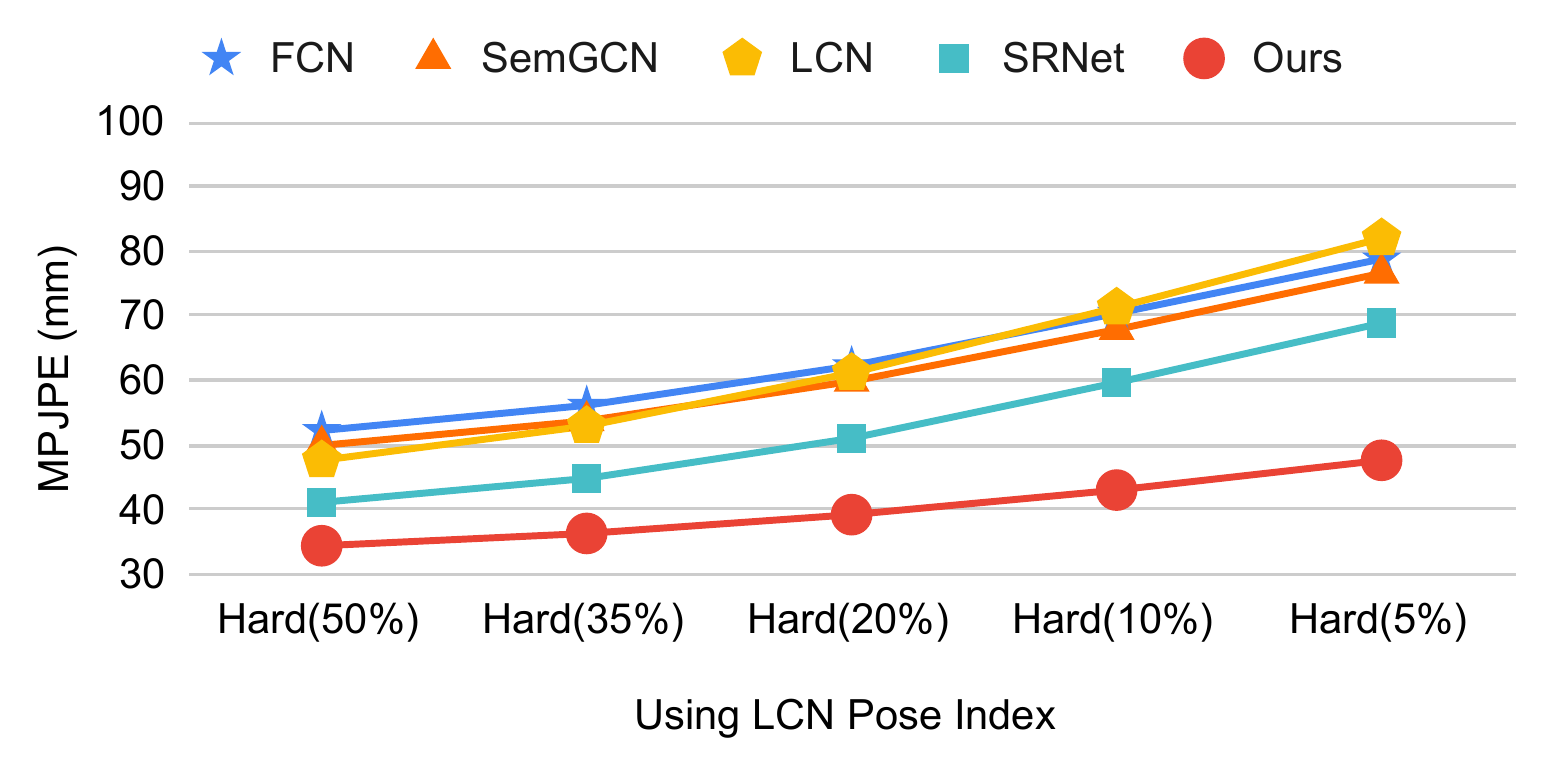}   
		\end{minipage}
	}
	\vspace{-0.3cm}
	\subfigure[The hard poses of SRNet~\cite{zeng2020srnet}] 
	{
		\begin{minipage}{8cm}
			\centering      
			\includegraphics[width=\linewidth]{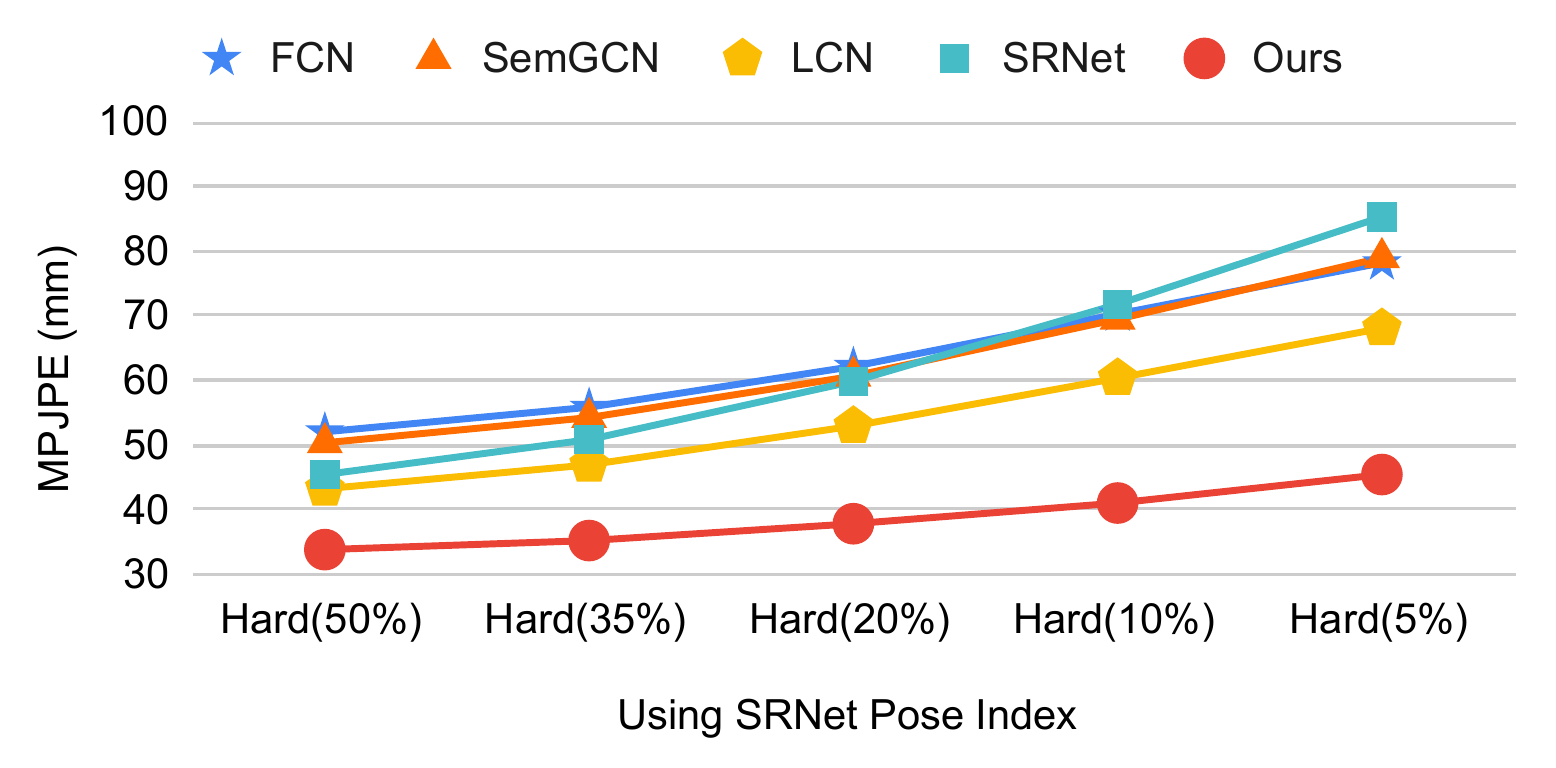}   
		\end{minipage}
	}
		\vspace{-0.3cm}
	\subfigure[The hard poses of Ours] 
	{
		\begin{minipage}{8cm}
			\centering      
			\includegraphics[width=\linewidth]{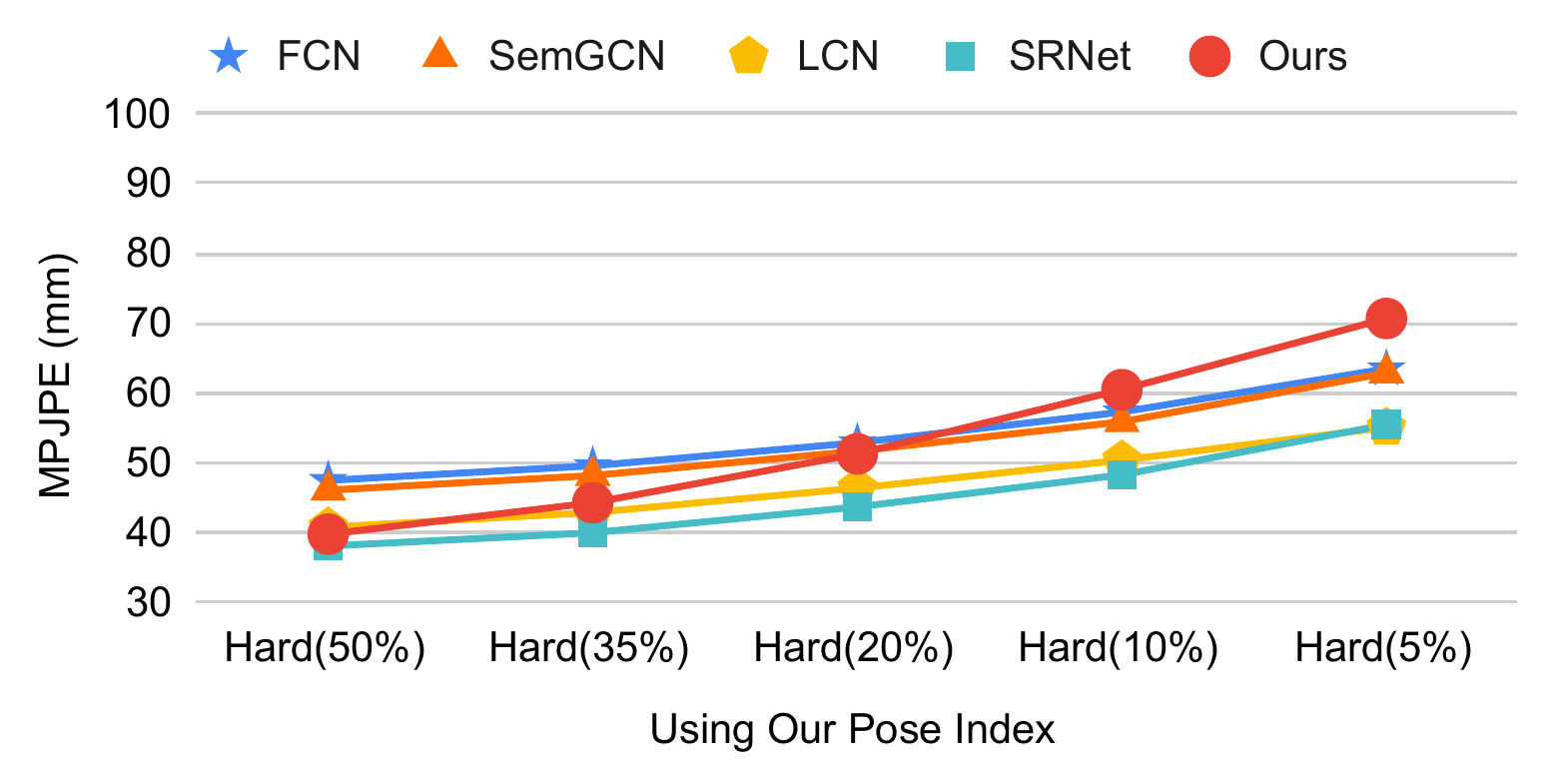}   
		\end{minipage}
	}
	\caption{The comparison of the hard poses in terms of each method.} 
	\label{fig:hard_result}
\end{figure}

\vspace{-0.2cm}
We define \emph{high-error poses} as hard poses in the 2D-3D pose regression task. After analyzing the error distribution of hard poses in recent works~\cite{martinez2017simple,zhao2019semantic,ci2019optimizing,zeng2020srnet}, we could conclude they are model-specific. As shown in Fig.~\ref{fig:hard_result}, we illustrate the comparison of the ($50\%\sim5\%$) hardest poses from each method. For example, Fig.~\ref{fig:hard_result}(a) shows the ($50\%\sim5\%$) hardest poses from the fully connected network~\cite{martinez2017simple}, and we compare the results with the other four methods under the same poses.

We can observe: (1) The hardest $10\%$ poses of each method is different, indicating that hard poses are model-specific; (2) as the poses become increasingly difficult, the errors of all methods rise to some extent; (3) our method obtains the best results for the hardest poses of all the other four methods; (4) the error gap in Fig.~\ref{fig:hard_result}(e) is smaller than Fig.~\ref{fig:hard_result}(a$\sim$d).

\subsection{Comparison in PA-MPJPE}
\label{sec:hm_pa}
\vspace{-0.2cm}
In Tab.~\ref{tab:human_p2_single}, we compare our methods with other related works using the PA-MPJPE metric where available. We show the results from different 2D inputs, using detected poses or ground truth poses. Our approach achieves the new state-of-the-art with different inputs. Specifically, we surpass ~\cite{zeng2020srnet} from 27.8mm to 24.8mm (relative 10.8\% improvement) with 2D ground truth input. Moreover, we improve upon ~\cite{liu2020comprehensive} from 41.2mm to 39.0mm (relative 5.3\% improvement) with 2D keypoint detection input. Our method can also show the superior in this metric, indicating the effectiveness of this method.

\begin{table*}
\begin{center}
\scriptsize
\resizebox{\textwidth}{!}
{\begin{tabular}{ l c c c c c c c c c c c c c c c c }
\hline
Method & Direct & Discuss & Eat & Greet & Phone& Photo & Pose  & Purcha. & Sit & SitD &Smoke &Wait &WalkD&Walk& WalkT & Avg.\\
\hline
Martinez et al.~\cite{martinez2017simple} & 39.5 &43.2& 46.4 &47.0 &51.0 &56.0 &41.4 &40.6& 56.5& 69.4 &49.2 &45.0 &49.5 &38.0 &43.1 &47.7  \\
Fang et al.~\cite{fang2018learning}&38.2 &41.7 &43.7 &44.9 &48.5 &55.3 &40.2 &38.2 &54.5 &64.4 &47.2& 44.3 &47.3 &36.7& 41.7& 45.7\\
Park et al.~\cite{park20183d}&38.3 &42.5& 41.5& 43.3 &47.5& 53.0& 39.3 &37.1& 54.1& 64.3& 46.0& 42.0& 44.8& 34.7& 38.7& 45.0\\
Hossain et al.~\cite{rayat2018exploiting} $\S$ & 35.7 &39.3 &44.6 &43.0 &47.2 &54.0 &38.3& 37.5& 51.6 &61.3 &46.5 &41.4& 47.3 &34.2 &39.4& 44.1\\
Zou et al.~\cite{zou2020high} $\dagger$&38.6 &42.8& 41.8 &43.4 &44.6& 52.9 &37.5 &38.6& 53.3& 60.0& 44.4 &40.9 &46.9 &32.2& 37.9& 43.7\\
Liu et al.~\cite{liu2020learning}$\dagger$ &38.4 &41.1 &40.6& 42.8& 43.5& 51.6 &39.5 &37.6 &49.7& 58.1& 43.2& 39.2& 45.2 &32.8 &38.1 &42.8\\
Ci et al.~\cite{ci2019optimizing}$\dagger$&36.9& 41.6& 38.0& 41.0& 41.9 &51.1 &38.2& 37.6& 49.1 &62.1 &43.1& 39.9& 43.5& 32.2& 37.0& 42.2\\
Liu et al.~\cite{liu2020comprehensive}$\dagger$&35.9 &40.0 &38.0 &41.5 &42.5 &51.4 &37.8 &36.0 &48.6 &56.6 &41.8& 38.3& 42.7& 31.7& 36.2 &41.2\\
\hline
\emph{Ours-HCSF}$\dagger$&34.3&37.6&37.5&38.6&39.5&44.2&38.3&35.5&48.5&55.6&41.4&38.7&42.3&30.8&32.2&39.7\\
\emph{Ours-HCSF w/A}$\dagger$&\textbf{33.9}&\textbf{37.2}&\textbf{36.8}&\textbf{38.1}&\textbf{38.7}&\textbf{43.5}&\textbf{37.8}&\textbf{35.0}&\textbf{47.2}&\textbf{53.8}&\textbf{40.7}&\textbf{38.3}&\textbf{41.8}&\textbf{30.1}&\textbf{31.4}&\textbf{39.0}\\

\hline
Zeng et al.~\cite{zeng2020srnet}$\S$&24.3&28.1&24.3& 28.1&27.4&29.8&28.3&25.6&27.8& 34.5&27.5&27.7&31.8&25.7&25.6&27.8\\
\emph{Ours-HCSF}$\dagger$ $\S$ &20.9&27.3&22.4&25.3&24.4&29.7&24.9&23.0&27.2&32.6&25.8&25.6&26.4&20.4&21.7&25.2\\
\emph{Ours-HCSF w/A}$\dagger$ $\S$ &\textbf{20.7}&\textbf{26.9}&\textbf{22.1}&\textbf{24.8}&\textbf{24.0}&\textbf{29.1}&\textbf{24.5}&\textbf{22.7}&\textbf{26.8}&\textbf{32.1}&\textbf{25.3}&\textbf{25.2}&\textbf{26.0}&\textbf{20.2}&\textbf{21.5}&\textbf{24.8}\\
\hline
\end{tabular}}
\end{center}
\vspace{-0.5cm}
\caption{Comparison results regarding PA-MPJPE after \emph{rigid transformation} from the ground truth. We highlight the graph-based methods by $\dagger$. $\S$ donates the use of 2D ground truth poses as input. Best results in bold.}
\vspace{-0.2cm}
\label{tab:human_p2_single}
\end{table*}

\subsection{Ablation Study on Dynamic Graph}
\vspace{-0.2cm}
\label{sec:hm_ab}
This work has two main contributions: Hierarchical Channel-Squeezing Fusion (HCSF) and temporal-aware dynamic graph learning. We further explore how temporal-aware dynamic graph alone influences the regression results. The 2D inputs are 2D ground truth to explore the upper bound of our method to avoid some irrelevant noises from detected 2D poses.

\noindent\textbf{Effects of dynamic graph learning.}
Dynamic graph learning shows different action-related connectivity with different inputs. It can be more flexible to extract specific-action patterns, especially for hard poses. We have demonstrated the influence on both HCSF and dynamic graph learning in the main paper. Accordingly, we study the effects of dynamic graph learning alone. We take the \emph{Non-hierarchy strategy} LCN with the static graph aggregating with hop-$2$ as a baseline. Similar to the Tab.$6$ in the main paper, the Tab. ~\ref{tab:hm_dy1}a, ~\ref{tab:hm_dy1}b, ~\ref{tab:hm_dy1}c shows that $\mM_k$ ($ori$), using the physical topology as an initial connections, is better than $\mM_k$ ($dense$) and $\mM_k$ ($rand$). The weighted graph $\mM_k$ ($ori$) can also surpass the \emph{same} weighted graph in LCN. Moreover, only learning graph structure from features increase the error from $35.7$mm to $46.1$, which is infeasible. After combining the weighed graph $\mM_k$ ($ori$) with the dynamic offset $\mO_k$, we can obtain $0.5$mm improvement. Furthermore, considering a dynamic scale $\alpha$ to control the influence of the dynamic offsets, which is the formula in Eq.$8$, will be helpful. Last, we can observe that the temporal-aware scheme can boost the performance, decreasing the MPJPE from $34.0$mm to $33.5$mm.

\begin{table*}[ht!]
\begin{center}
\small
{
\begin{tabular}{ l| c |c |c |c |c|c|c|c}
\hline
Method & LCN & a & b & c & d & e & f & g\\
\hline
$\mA_k$ & $ori$ &Only $\mM_k$ ($ori$) & Only $\mM_k$ ($dense$)&Only $\mM_k$ ($rand$) & Only $\mO_k$ & $\mM_k + \mO_k$ & Eq.$8$ & Eq.$8$ w/T\\
\hline\hline
MPJPE(mm)&35.7&34.8&35.5&41.2&46.1&34.3&34.0&\textbf{33.5}\\
\hline
\end{tabular}}
\end{center}
\vspace{-0.5cm}
\caption{Comparison on the effects of dynamic graph learning $\mA$ in a \emph{Non-hierarchy strategy}. $ori$ is the static graph with physical connections, shown in LCN~\cite{ci2019optimizing}. \emph{Baseline} takes $\mA_k$ as $ori$. \emph{Only $\mM_k$ ($\cdot$)} denotes applying $\mM_k$ with different initialization. \emph{Only $\mO_k$} keeps the dynamic offset in Eq.$8$. \emph{$\mM_k + \mO_k$} equals to set $\alpha=1$ in Eq.$8$. \emph{w/T} represents the temporal-aware scheme defined in Sec.$3.3$.}
\label{tab:hm_dy1}
\end{table*}

\noindent\textbf{Effects of the temporal scale.}
The uncertainty in single-frame poses will affect the regression results, making dynamic graph learning unstable and misleading. Hence, it is essential to introduce temporal consistency to make the process effective. We then explore how different settings in the temporal-aware scheme impact the performance. The temporal-aware schemes are different from the receptive fields. We fix $S$=$1$, $L$=$2$, $d$=$1/8$. The channel size of each layer is $128$. And the frame of input is $9$. From Tab.~\ref{tab:hm_tp}, we can find that using the $3\times1$ kernel size will be better than other settings. And using temporal information will consistently improve the single-frame results by $0.1\sim0.4$mm. Thus, we report our final results using the $3\times1$ kernel size.

\begin{table}
\begin{center}
\scriptsize
{
\begin{tabular}{ l| c |c |c |c|c|c}
\hline
$F$ &(1,1)&(3,1)&(3,1) w/$st.$=2&(3,1) w/$di.$=2&(5,1)&(7,1)\\
\hline
HCSF&30.8&\textbf{30.4}&30.7&30.7&30.6&30.7\\
\hline
\end{tabular}}
\end{center}
\vspace{-0.5cm}
\caption{The impact of settings $F$ of temporal convolution in dynamic graph learning of 3D human pose estimation. $st.$ is an abbreviation for $stride$, and $di.$ is $dilation$. 
}
\label{tab:hm_tp}
\end{table}

\section{Skeleton-based Human Action Recognition}
\vspace{-0.2cm}
 In this section, we present the experimental details, more results and ablation study of skeleton-based action recognition in Sec.~\ref{sec:act_imp}, Sec.~\ref{sec:har_single} and Sec.~\ref{sec:act_ab}, respectively.
\subsection{Dataset and Implementation Details} 
\label{sec:act_imp}
\subsubsection{Data Description}
\vspace{-0.2cm}
\noindent\textbf{NTU RGB+D 60}~\cite{shahroudy2016ntu} is one of the most widely used in-door RGB+Depth action recognition dataset with 60 actions. They include daily, mutual, and health-related actions. NTU RGB+D 60 has 40 subjects under three cameras.
Following~\cite{shi2019skeleton,shi2020skeleton,ye2020dynamic,peng2020mix,yan2018spatial}, we use skeleton sequences with 25 body joints captured by Kinect V.2 as inputs, and take two evaluation settings in NTU RGB+D 60: (1) Cross-Subject (X-Sub), where 20 subjects each for training and testing, respectively; (2) Cross-View (X-View), where 2 camera views for training and 1 camera view for testing. We perform the ablation study in Sec.~\ref{sec:act_ab} on the X-View setting.

\noindent\textbf{NTU RGB+D 120}~\cite{liu2019ntu} collects 120 various actions by 106 distinct subjects and contains more than 114 thousand video samples and 8 million frames. We also follow some previous works~\cite{ye2020dynamic,liu2020disentangling,peng2020mix,peng2020learning}, using two evaluation settings: (1) Cross-Setup (X-Set), training on 16 camera setups and testing on other 16 camera setups; (2) Cross-Subject (X-Sub), half subjects for training and half for testing. 
We report the top-1 accuracy on both benchmarks.

\subsubsection{Data Pre-processing} 
\vspace{-0.2cm}
The procedure for both datasets follows~\cite{shi2019skeleton,shi2020skeleton,liu2020disentangling}. Each video has a maximum of 300 frames, and if it is shorter than 300, we repeat some frames to make up for it. Since there are at most two people in both datasets, we pad the second body with zeros to keep the same shape of inputs when the second body does not appear. 
\vspace{-0.2cm}

\subsubsection{Training Details} 
\vspace{-0.2cm}

We build a ten-layer network, including nine cascaded blocks that consist of one HCSF layer followed by BN, ReLU, temporal convolution layer (TCN), BN and ReLU. Each temporal 1D convolution layer conducts $9 \times 1$ convolution on the feature maps. Each block is wrapped with a residual connection. The output dimension for each block are 64, 64, 64, 128, 128, 128, 256, 256 and 256. A global average pooling layer and a fully-connected layer are used to aggregate extracted features, and then, feed them into a softmax classifier to obtain the action class. The above framework is also a common setting as in~\cite{yan2018spatial,shi2019skeleton,shi2020skeleton,ye2020dynamic}. For multi-stream networks~\cite{shi2020skeleton}, we use four modalities, e.g., joints, bones and their motions, as inputs for each stream, and average their softmax scores to obtain the final prediction. Cross-entropy is used as the classification loss function to back-propagate gradients. We set the entry values in the adjacency matrix to be ones if two nodes are physically connected and zero if not. 

For the training settings, we train our model for $60$ epochs using the SGD optimizer with mini-batch size $64$. The initial learning rate is 0.1 and it reduces by $10$ times in both the $35_{th}$ and $45_{th}$ epoch, respectively. The weight decay is set as $0.0005$. All data augmentation is the same as~\cite{shi2019skeleton, shi2020skeleton}.

\subsection{Results of Single-Stream Framework}
\label{sec:har_single}
Due to space limitations, we only report the accuracy of the multi-stream framework~\cite{shi2020skeleton} for the skeleton-based human action recognition task in the main paper. Specifically, the multi-stream network comprises four different modality inputs: the 3D skeleton joint position, the 3D skeleton bone vector, the motion of the 3D skeleton joint, and the motion of the 3D skeleton bone. Here, we report the performance of \emph{each modality input} in Tab.~\ref{tab:har_stream} for the ease of comparison with existing works.

\begin{table}
\begin{center}
\scriptsize
{
\begin{tabular}{ c| cc|c c }
\hline
\multirow{2}{*}{Method}& \multicolumn{2}{c|}{\textbf{NTU-RGB+D 60}} &  \multicolumn{2}{c}{\textbf{NTU-RGB+D 120}} \\
&X-Sub(\%)&X-View(\%)&X-Sub(\%)&X-Set(\%)\\
\hline
Joint&89.0&95.3&83.5&85.7\\
Bone&89.3&94.9&85.0&86.6\\
Joint-Motion&86.9&93.5&80.1&81.5\\
Bone-Motion&86.9&93.1&80.6&83.0\\
\hline
Multi-Stream&\textbf{91.6}&\textbf{96.7}&\textbf{87.5}&\textbf{89.2}\\
\hline

\end{tabular}}
\end{center}
\vspace{-0.5cm}
\caption{Top-1 accuracy (\%) is used as the evaluation metric. The best result in each K is in bold.
}
\label{tab:har_stream}
\end{table}

\subsection{Ablation Study}
\label{sec:act_ab}
We investigate the proposed methods on the NTU RGB-D X-View setting with 3D joint positions as inputs.

\paragraph{\textbf{Effects of hierarchical channel-squeezing fusion block.}}
From Tab.~\ref{tab:decay_D}, our method improves the accuracy of 0.7$\%$ steadily under all three graph settings, static graphs $\mathcal{G}_k$ and two dynamic graphs $\mathcal{M}_k$ and $\mathcal{A}_{k}$ in Eq.$8$. Basically, better results can be achieved when $d$=$1/8$.
Moreover, we get the best results when using HCSF with dynamic graph $\mathcal{A}_{k}$, which validates the effectiveness of the proposed structure.

\begin{table}
\begin{center}
\scriptsize
{
\begin{tabular}{ l| c |c |c |c|c}
\hline
Decay Rate $d$&$1$&$1/2$&$1/4$&$1/8$&$1/16$\\
\hline

Static-$\mathcal{G}$& 93.9&94.5&94.6&\textbf{94.8}&94.5\\
Dynamic-$\mathcal{M}$ &94.4&94.9&\textbf{95.1}&94.9&\textbf{95.1}\\
Dynamic-$\mathcal{A}$ &94.6&95.0&95.2&\textbf{95.3}&\textbf{95.3} \\
\hline
\end{tabular}}
\end{center}
\vspace{-0.5cm}
\caption{The impact of decay rate $d$ under static matrix $\mathcal{G}$, dynamic graph from $\mathbf{M_k}$, and dynamic graph from $\mathbf{A_{k}}$ in Eq.$8$.
}
\label{tab:decay_D}
\vspace{-10pt}
\end{table}

\vspace{5pt}
Furthermore, in Tab.~\ref{tab:comp_dy}, we demonstrate the performance of different methods concerning the number of hops. Since the skeleton topology in NTU-RGBD datasets is different from Human3.6M, it has more keypoints and further hops. The furthest hop is $13$ in our pre-defined topology. We set $S$=$5$, $L$=$7$ and $d$=$1/8$. $k$-hop ($k$=$1, 5, 7$) means aggregating the neighbors within the distance $k$ (1-hop with a static graph is ST-GCN~\cite{yan2018spatial}). Mixhop~\cite{abu2019mixhop} means that it concatenates the $k$-hop ($k$=$1, 5, 7$) features as the output of a layer, and the output size of the $k$-hop feature is one-third of the final output. MS-Hop means that it averages the $k$-hop ($k$=$1, 5, 7$) features, and the output size of the $k$-hop feature is the same as the final output. 

As illustrated in Tab.~\ref{tab:comp_dy}, though MixHop and MS-Hop show improvements on k-hop strategies, they have no distinction in handling distant and close neighbors, which over-mix the useful and noisy information.
Our approaches outperform all other baselines, which indicates the effectiveness of the hierarchical channel-squeezing fusion strategy.

Additionally, we explore the effects of other hyper-parameters in the HCSF. We have the following observations. First, when using a dynamic graph $\mathcal{A}_{k}$ in Eq.$8$ and fixing the hyper-parameters squeezing ratio $d$ and the output channel size $C$ in a layer, we find little effects on the results that $S$ and $L$ has. The accuracy is stable around 95.1$\%$ $(\sim0.2\%)$. It indicates that the HCSF is robust to the noise in the graph.
Second, as the number of hops increases, the performance first improves and then becomes stable. Since adding more hops leads to extra computations, to balance the computation efficiency and performance, our final setting for each layer is $S$=$5$, $L$=$7$, $d$=$1/8$, $C$ of each layer is the same as~\cite{yan2018spatial,shi2019skeleton,shi2020skeleton}. 
Last, we also explore to automatically learn the relations between hops and dimensions with the guidance of channel attention. However, we find that the exponentially decaying in dimension consistently yields better results than the soft attention, which may be because the soft attention mechanism introduces more uncertainty and complexity.

\begin{table}
\begin{center}
\scriptsize
{
\begin{tabular}{ l| c |c |c |c|c |c}
\hline
Method&1-hop&5-hop&7-hop&MixHop&MS-Hop&Ours\\
\hline

Static $\mathcal{G}$& 92.2&93.5&93.7&93.9&94.1&\textbf{94.8}\\
Dynamic-$\mathcal{M}$ &93.4&94.1&94.1&94.5&94.6&\textbf{95.2}\\
Dynamic-$\mathcal{A}$&93.9&94.3&94.2&94.8&94.7&\textbf{95.3} \\
\hline
\end{tabular}}
\end{center}
\vspace{-0.5cm}

\caption{Comparison on various multiple hop structures under static matrix $\mathcal{G}_K$, dynamic graph from $\mathcal{M}_k$, and a dynamic graph from $\mathcal{A}_{k}$. Top-1 accuracy is used as the evaluation metric.
}
\label{tab:comp_dy}
\end{table}

\vspace{-0.3cm}
\paragraph{\textbf{Effects of the temporal-aware dynamic graph learning.}}

The jitter and missing inputs will make dynamic graph learning unreliable, making it difficult to distinguish between similar actions,  e.g., ``eat a meal" and  ``brushing teeth." Such problems are serious in using single-frame features, but they can be improved by involving temporal information. From Tab.~\ref{tab:har_temporal}, we can observe that when using three frames into a temporal convolution, it can improve the single-frame setting by 0.6\%. While the settings of temporal aggregation are important, the longer temporal contexts will also degrade the performance, and use three frames will be the optimal setting.

\begin{table}
\begin{center}
\scriptsize
{
\begin{tabular}{ l| c |c |c |c|c|c}
\hline
$F$ &(1,1)&(3,1)&(3,1) w/$st.$=2&(3,1) w/$di.$=2&(5,1)&(7,1)\\
\hline
HCSF&94.7&\textbf{95.3}&95.0&94.8&95.1&94.7\\
\hline
\end{tabular}}
\end{center}
\vspace{-0.5cm}
\caption{The impact of settings of temporal convolution in dynamic graph learning of skeleton-based action recognition. $st.$ is an abbreviation for $stride$, and $di.$ is $dilation$.
}
\label{tab:har_temporal}
\vspace{-10pt}
\end{table}

\end{document}